\documentclass[conference]{IEEEtran}

\usepackage{graphicx}
\usepackage{bm}
\usepackage{amsmath}
\usepackage{amssymb}
\usepackage{esvect}
\usepackage{amsmath}
\usepackage{amssymb}

  \newcommand{\T}{^\mathsf{T}}
  
  \newcommand{\R}{\mathbb{R}}

  \newcommand{\N}{\mathrm{N}}

  \newcommand{\diff}{\,\mathrm{d}}

  \usepackage{bm}
  
  \newcommand{\mbf}[1]{\mathbf{#1}}
  \newcommand{\vect}[1]{\mbf{#1}}
  \newcommand{\vectb}[1]{\bm{#1}}

  \newcommand{\eg}{\textit{e.g.}}
  \newcommand{\ie}{\textit{i.e.}}

  \newcommand{\etal}{\textit{et~al.}}

  \usepackage{booktabs}

  \usepackage{tikz,pgfplots}
  \usetikzlibrary{plotmarks,shapes,arrows}
  \pgfplotsset{compat=newest} 
  \pgfkeys{/pgf/number format/.cd,1000 sep={}}

  \newlength\figureheight
  \newlength\figurewidth

  \newcommand{\url}[1]{#1}

  \usetikzlibrary{external}

  \makeatletter
  \let\NAT@parse\undefined
  \makeatother

  \makeatletter
  \let\NAT@parse\undefined
  \makeatother
  \usepackage[square,numbers,sort&compress]{natbib}

  \makeatletter
  \let\MYcaption\@makecaption
  \makeatother

  \usepackage[font=footnotesize]{subcaption}

  \makeatletter
  \let\@makecaption\MYcaption
  \makeatother

\begin{document}

\title{Robust Gyroscope-Aided Camera Self-Calibration}

\author{\IEEEauthorblockN{Santiago Cort\'{e}s Reina}
\IEEEauthorblockA{Department of Computer Science\\
Aalto University\\
Helsinki, Finland\\
santiago.cortesreina@aalto.fi}
\and
\IEEEauthorblockN{Arno Solin}
\IEEEauthorblockA{Department of Computer Science\\
Aalto University\\
Helsinki, Finland\\
arno.solin@aalto.fi}
\and
\IEEEauthorblockN{Juho Kannala}
\IEEEauthorblockA{Department of Computer Science\\
Aalto University\\
Helsinki, Finland\\
juho.kannala@aalto.fi}
}

\maketitle

\maketitle

\begin{abstract}
  Camera calibration for estimating the intrinsic parameters and lens distortion is a prerequisite for various monocular vision applications including feature tracking and video stabilization. This application paper proposes a model for estimating the parameters on the fly by fusing gyroscope and camera data, both readily available in modern day smartphones. The model is based on joint estimation of visual feature positions, camera parameters, and the camera pose, the movement of which is assumed to follow the movement predicted by the gyroscope. Our model assumes the camera movement to be free, but continuous and differentiable, and individual features are assumed to stay stationary. The estimation is performed online using an extended Kalman filter, and it is shown to outperform existing methods in robustness and insensitivity to initialization. We demonstrate the method using simulated data and empirical data from an iPad.
\end{abstract}

\section{Introduction}
\noindent
The growth in the market of smartphones and tablets has brought monocular cameras to even the cheapest of smart-devices. Simultaneously, the improved computational capabilities have made it possible to expand their use into new fields and use cases, such as video calls, payment verification, and augmented reality. However, employing the device camera in pose estimation, video stabilization, or feature tracking requires the camera calibration parameters to be known or estimated.

On smart-devices, the use cases often provide an explicit calibration step that includes capturing a pre-defined calibration pattern from different positions. This procedure is the base for traditional camera calibration (see, \eg, \cite{Hartley+Zisserman:2004} for an overview). Self-calibration is the problem of modeling the internal parameters of a camera (projection matrix and distortion coefficients) without using any known pattern. Luckily there are usually additional sensors available on the devices, typically a low-cost MEMS inertial measurement unit (IMU). The IMU typically provides fast-sampled (up to some hundreds of Hz) readings of the specific force (accelerometer) and turn-rate (gyroscope) in the device's coordinate frame (see, \eg, \cite{Solin+Cortes+Rahtu+Kannala:2018-FUSION} for a more thorough introduction).

In this work we propose a method for camera self-calibration using the information from a gyroscope rigidly attached to the camera. We use a structure from motion (SfM) approach, where the camera model minimizes the reprojection error over a set of images. The proposed method jointly estimates the camera pose, tracked feature positions, and camera parameters. The method works online employing a state-space formulation, where the fast-sampled gyroscope data is used for forward-predicting the relative camera and feature movement, and the visual data tracking results are then matched to the predictions in a probabilistic fashion. The inference itself is solved by an extended Kalman filter.

\begin{figure}[t!]
  \centering\footnotesize
  \begin{tikzpicture}
    \node[anchor=south west,inner sep=0] (image) at (0,0) %
      {\includegraphics[width=\columnwidth]{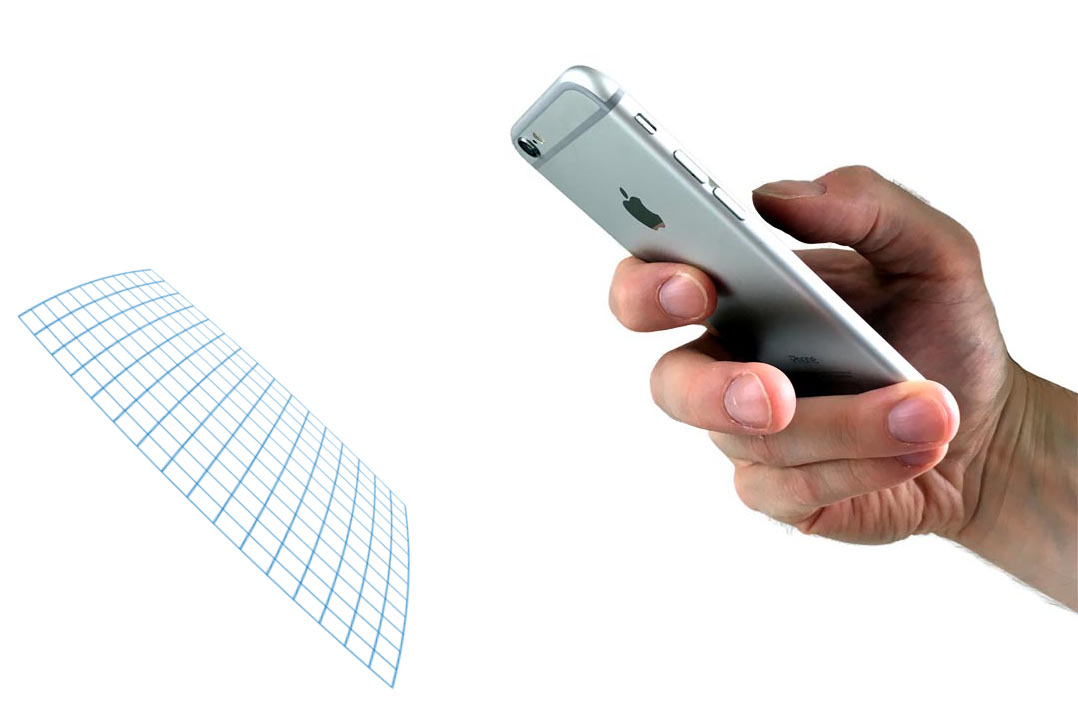}};
    \begin{scope}[x={(image.south east)},y={(image.north west)}]

      \tikzstyle{label} = [align=center]
      \tikzstyle{line} = [draw]
      \tikzstyle{point} = []

      \node [label] (lab_gyro)       at (0.75,0.95) {Phone \\ gyroscope \\ with turn rate  \\ $\vectb{\omega}$ [rad/s]};
      \node [label] (lab_cam)        at (0.25,0.95) {Phone camera \\ position $\vect{p}$ \\ orientation $\vect{q}$};
      \node [label] (lab_dist)       at (0.55,0.10) {Radial distortion \\ ($k_1$ and $k_2$)};
      \node [label] (lab_param)      at (0.20,0.20) {Camera parameters \\ ($f_{x/y}$ and $c_{x/y}$)};

      \node [point] (gyro)           at (0.60,0.75) {};
      \node [point] (camera)         at (0.4879, 0.7877) {};
      \node [point] (sw)             at (0.0158, 0.5484) {};
      \node [point] (nw)             at (0.3636, 0.0214) {};
      \node [point] (se)             at (0.3757, 0.2821) {};
      \node [point] (ne)             at (0.1401, 0.6154) {};

      \path [line] (lab_gyro) |- (gyro);
      \path [line] (lab_cam)  -| (camera);
      \path [line] (lab_cam)  -| (camera);

      \draw [densely dotted,  thin] (camera) -- (sw);
      \draw [densely dotted,  thin] (camera) -- (se);
      \draw [densely dotted,  thin] (camera) -- (nw);
      \draw [densely dotted,  thin] (camera) -- (ne);

    \end{scope}
  \end{tikzpicture}
  \caption{Illustration of the model setup: A pinhole camera with position $\vect{p}$ and orientation quaternion $\vect{q}$, auxiliary turn rate $\vectb{\omega}$ from the gyroscope. A set of (unknown) feature locations $\{\vect{z}_i\}$ are observed through a camera model with linear and non-linear distortion parameters.}
  \label{fig:camera-model}
\end{figure}
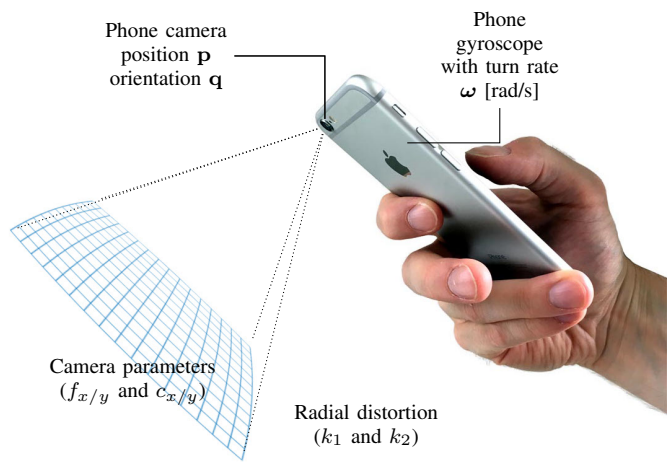

Due to its practical importance, camera self-calibration---or auto-calibration---has been studied extensively during the past decades. Outside smartphones, practical applications are found, for example, in traffic surveillance \cite{alvarez+llorca:2014}, projector camera calibration \cite{Li+Sekkati:2017,WIlli+grundhofer:2017} and robotics for autonomous vehicles \cite{Sun+Wang:2017,Gopaul+Wang:2016,Maye+Furgale:2013,Goldshtein+Oshman:2007,Faion+Ruoff:2012,Heidari+Alaei-Novin:2013}. The seminal paper by Faugeras~\etal~\cite{Faugeras+Luong+Maybank:1992} introduced the concept of calibrating the intrinsic and extrinsic parameters without a known calibration object or pattern. Since then many methods have been introduced, building upon special kind of motion (\eg\ purely rotating or planar), special scene geometry (\eg\ planar scenes or special depth structure, see \cite{Herrera+Kannala+Heikkila:2016} for discussion), or auxiliary sensors.

For example, Civera~\etal\ \cite{Civera+Bueno+Davison+Montiel:2009} proposed a sum-of-Gaussians filter approach building upon a SfM approach where there has to be sufficient translation. Additional sensors ease the motion constraints. Using both a gyroscope and an accelerometer for calibration, information can be acquired  about the relative rotation, absolute scale, and the world coordinate frame orientation (see, \eg, \cite{Li+Yu+Zheng+Mourikis:2014}). Similar setups are often employed in visual-inertial odometry methods (see, \eg, the discussion in \cite{Shelley:2014,Solin+Cortes+Rahtu+Kannala:2018-WACV}), where the same set of sensors are used and the camera calibration is estimated as a part of the inference. Gyroscopes are also used in video stabilization, and \cite{Ovren+Forssen:2015} proposed a model for scaling, time offset, and relative pose calibration using the gyroscope. They, however, do not estimate the camera calibration parameters.

There are also other methods which only use the camera and a gyroscope. Karpenko~\etal\ \cite{Karpenko+Jacobs+Baek+Levoy:2011} proposed a method for calibration which uses a gyroscope together with a quick shake when pointing to far-away objects, while Hwangbo~\etal\ \cite{Hwangbo+Kim+Kanade:2011} constrain the estimation by assuming pure rotation.

Within methods designed for fusing camera and gyroscope data, we are only aware of one previously published approach, which works in the same setting as the proposed method. The method by Jia and Evans was first published in \cite{Jia+Evans:2013} and later refined in \cite{Jia+Evans:2014}. Their method together with ours does not assume special movement and only require visual and gyroscope data.

The proposed model is based on the following assumptions: (i)~The camera movement is free, but continuous and differentiable (rather smooth), (ii)~The camera rotations follow the gyroscope turn rate observations, (iii)~The world coordinates of individual features are unknown, but assumed constant (individual features assumed to stay stationary). In comaprison to \cite{Jia+Evans:2014}, this method aims at providing higher robustness to initialization and feature-poor enivironments with only a few visual features being tracked. Where \cite{Jia+Evans:2014} uses a lot of short (two-frame) connections of features, our method uses a few long chains of connected features.

This paper is structured as follows. Section~\ref{sec:methods} presents the theoretical methodology in detail starting from the camera calibration model, the proposed state-space model, and finally how to jointly infer all the unknown parameters. The Results section shows the method employed both in a simulation study and a on real-world data. Finally, the assumptions and modeling problems and some future work are discussed.

\section{Methods}
\label{sec:methods}
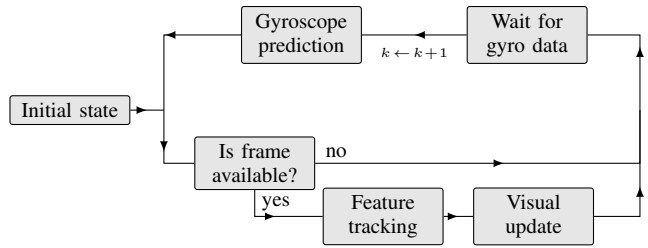
\begin{figure}
  \centering\footnotesize

  \tikzstyle{decision} = [draw, node distance=1cm, rounded corners=1pt, draw, fill=black!10, text width=5em, text badly centered]
  \tikzstyle{block} = [draw, node distance=1cm, text badly centered, rounded corners=1pt, text width=5em, fill=black!10]
  \tikzstyle{node} = [inner sep=0, outer sep=0, minimum size=0]
  \tikzstyle{line} = [draw, postaction={decorate}]

  \usetikzlibrary{decorations.markings}

  \begin{tikzpicture}[node distance=2cm, auto, decoration={markings,mark=at position 0.5 with {\arrow{latex}}}]

    \node[block] (init) {Initial state};
    \node[node, right of=init, xshift=-0.75cm] (start_node) {};
    \node[decision, below right of=start_node, xshift=0.5cm] (choose) {Is frame available?};
    \node[block, below right of=choose, xshift=1cm] (track) {Feature tracking};
    \node[block, right of=track, xshift=1cm] (update) {Visual update};
    \node[node, above right of=update] (advance) {};
    \node[block, above of=advance, xshift=-1.5cm] (wait) {Wait for gyro data};
    \node[block, left of=wait, xshift=-2cm] (predict) {Gyroscope prediction};

    \path [line] (init) -- (start_node);
    \path [line] (start_node) |- (choose);
    \path [line] (choose) |-node[anchor=south,xshift=1em,yshift=-0.1em] {yes} (track);
    \path [line] (track) -- (update);
    \path [line] (update) -| (advance);
    \path [line] (choose) node[anchor=west,xshift=3em,yshift=0.5em] {no}-| (advance);
    \path [line] (advance) |- (wait);
    \path [line] (wait) --node[anchor=south,xshift=0em,yshift=-1.5em] {\tiny$k\!\leftarrow\!k\!+\!1$} (predict);
    \path [line] (predict) -| (start_node);

  \end{tikzpicture}
  \caption{The information flow in the self-calibration method. The gyroscope prediction loop runs at 100~Hz, and visual updates occur once a new frame is acquired (typically at 10 Hz).}
  \label{fig:flow}
\end{figure}

\noindent
We start by defining the notation used in the camera calibration model, which will later be used in specification of the measurement model. Then we proceed to setting up the state-space model for the gyroscope-aided dynamics of the camera pose, feature positions, and camera parameters. Finally, we couple the forward dynamics with the image data in a visual measurement model. A sketch of the information flow in the method is shown in Figure~\ref{fig:flow}.

\subsection{Camera Model}
\noindent
We build upon the well-established theory of monocular camera calibration models. A detailed and extensive description of camera models can be found in \cite{Hartley+Zisserman:2004}. Through the camera model, points $(x,y,z)$ in three-dimensional space are projected to image plane coordinates $(u,v)$ using a pinhole camera model. First the points are rigidly transformed into the camera reference frame, that is origin at the optical center, the $z$-axis perpendicular to the image plane and $y$-axis parallel to the vertical axis of the image. Then the points are projected into the image plane using the pinhole projection. In matrix form
\begin{equation}
  \begin{pmatrix}
    u \\
    v \\
    1\\
  \end{pmatrix} 
  = \vect{K} \, \vect{E} \,
  \begin{pmatrix}
  	x\\
    y\\
    z\\
    1\\
  \end{pmatrix},
  \label{eq:pinhole1} 
\end{equation}
where the intrinsic matrix $\vect{K} \in \R^{3\times3}$ and extrinsic matrix $\vect{E}\in \R^{3\times4}$ are parametrized as follows:
\begin{equation}
  \vect{K} =\begin{pmatrix}
    f_x & 0 & c_x \\
    0 & f_y & c_y \\
    0 & 0 & 1 \\
  \end{pmatrix}
  \quad \text{and} \quad 
  \vect{E} =\begin{pmatrix}
    {\vect{R}\T} & -{\vect{R}\T}\vect{p} 
  \end{pmatrix}.
   \label{eq:pinhole2} 
\end{equation}
The parameters are the focal length ($f_x$ and $f_y$, separate between dimensions to account for non-square pixels), the origin of the image plane $(c_x, c_y)$, and the position $\vect{p} \in \R^3$ and orientation $\vect{R} \in \R^{3\times3}$ of the camera in the global frame of reference.

The projection operation is entirely linear (in homogeneous space), but real-world lenses usually introduce non-linear mappings. These are known as distortions, the most common distortion is rotationally symmetric around the image center and is modeled by so-called radial distortion coefficients $k_1$ and $k_2$ as follows:
\begin{equation}
  \begin{pmatrix}
     u' \\ v'
  \end{pmatrix}
  =
  \begin{pmatrix}
     u \\ v 
  \end{pmatrix} \, 
  (1 + k_1 \, r^2 + k_2 \, r^4),
  \label{eq:radial1}
\end{equation}
where the radial component is determined by
\begin{equation}
  r = \sqrt{\Big(\frac{u-c_x}{f_x}\Big)^2+\Big(\frac{v-c_y}{f_y}\Big)^2}. \label{eq:radial2}
\end{equation}

This camera model is shown in the illustrative sketch in Figure~\ref{fig:camera-model}, where also the resulting distortion effect is visible. We refer to the non-linear distortion function as `$\mathrm{distort}$' later in the paper.

\subsection{State Estimation}
\noindent
In order to include time-dependency between observed frames and rotational information from the gyroscope, we define a state-space model. The model describes a device with a monocular camera and a rigidly attached gyroscope with a known relative orientation between them (see Fig.~\ref{fig:camera-model}). An example of such a device is a modern smartphone. The following section shows the non-linear filer designed for information fusion, the non-linear filter allows for estimation and fusion while keeping track of the uncertainty and dependencies between the non-deterministic processes describing the evolution of the calibration parameters, camera pose, and feature locations.

We define a state-space model (see, \eg, \cite{Sarkka:2013}), where the state vector is 
\begin{equation}
  \vect{x} =
  \begin{pmatrix}
    \vect{c}\T &
    \vect{p}\T &
    \vect{v}\T &
    \vect{q}\T &
    \vect{z}\T
  \end{pmatrix} \T.
\end{equation}
The variable $\vect{c} = (f_x, f_y, c_x, c_y, k_1, k_2)$ contains the internal camera parameters, $\vect{p} \in \R^{3}$ and $\vect{v} \in \R^{3}$ contain the position and velocity of the camera, $\vect{q}$ contains the quaternion encoding the orientation of the camera, and $\vect{z} \in \R^{3d}$ contains the locations of the features ($d$ is the number of features being tracked).

The non-linear state-space model with the auxiliary gyroscope control signal $\vectb{\omega}_k$ is given as follows. The control can be embedded into the time-varying dynamical model such that $\vect{f}_k(\vect{x},\vectb{\varepsilon}_k) := \vect{f}(\vect{x},\vectb{\omega}_k,\vectb{\varepsilon}_k)$. The state-space model takes the canonical form:
\begin{equation}
\begin{aligned}
  \vect{x}_k &= \vect{f}_k(\vect{x}_{k-1} ,  \vectb{\varepsilon}_k), \\
  \vect{y}_k &= \vect{h}_k(\vect{x}_k) + \vectb{\gamma}_k,
\end{aligned}
\end{equation}
where $\vect{x}_k \in \R^n$ is the state at time step $t_k$, $k=1,2,\ldots$, $\vect{y}_k \in \R^m$ is a measurement, $\vectb{\varepsilon}_k \sim \N(\vectb{0}, \vect{Q}_k)$ is the Gaussian process noise, and  $\vectb{\gamma}_k \sim \N(\vectb{0},\vectb{\Sigma}_k)$ is the Gaussian measurement noise. The dynamics and measurements are specified in terms of the dynamical model function $\vect{f}(\cdot, \cdot, \cdot)$ and the measurement model function $\vect{h}_k(\cdot)$.

In this work we employ the extended Kalman filter (EKF, \cite{Jazwinski:1970, Bar-Shalom+Li+Kirubarajan:2001}) which provides a means of approximating the state distributions $p(\vect{x} \mid \vect{y}_{1:k})$ with Gaussians:
\begin{equation}
  p(\vect{x}_k \mid \vect{y}_{1:k}) \simeq \N(\vect{x}_k \mid \vect{m}_{k \mid k}, \vect{P}_{k \mid k}).
\end{equation}
In the EKF, these approximations are formed by first-order linearizations of the non-linearities. The extended Kalman filtering recursion can be written as follows (see \cite{Sarkka:2013} for detailed presentation). The dynamics are incorporated into the {\em prediction step}:
\begin{equation}
\begin{split}
  \vect{m}_{k \mid k-1} &= \vect{f}_k(\vect{m}_{k-1 \mid k-1}, \vectb{0}), \\
  \vect{P}_{k \mid k-1} &= \vect{F}_\vect{x}(\vect{m}_{k-1 \mid k-1}) \, \vect{P}_{k-1 \mid k-1} \, \vect{F}_\vect{x}\T(\vect{m}_{k-1 \mid k-1}) + \\ &\qquad
   \vect{F}_{\vectb{\varepsilon}}(\vect{m}_{k-1 \mid k-1}) \, \vect{Q}_k \, \vect{F}_{\vectb{\varepsilon}}\T(\vect{m}_{k-1 \mid k-1}),
\end{split}
\end{equation}
where the dynamic model is evaluated with the outcome from the previous step and zero noise, and $\vect{F}_\vect{x}(\cdot)$ denotes the Jacobian matrix of $\vect{f}_k(\cdot, \cdot)$ with respect to $\vect{x}$ and $\vect{F}_{\vectb{\varepsilon}}(\cdot)$ with respect to the process noise $\vectb{\varepsilon}$.

We will also address the special case, where the dynamics are entirely linear, that is $\vect{f}_k(\vect{x}) = \vect{A}_k$. In that case the filter prediction step is entirely given by the standard Kalman filter preduction step:
\begin{equation}
\begin{split}
  \vect{m}_{k \mid k-1} &= \vect{A}_k \, \vect{m}_{k-1 \mid k-1}, \\
  \vect{P}_{k \mid k-1} &= \vect{A}_k \, \vect{P}_{k-1 \mid k-1}\,\vect{A}_k\T + \vect{Q}_k.
\end{split}
\end{equation}

Measurement data providing observations of the system state at given time steps are combined with the model in the {\em update step}:
\begin{equation}
\begin{split} \label{eq:update}
  \vect{v}_k &= \vect{y}_k - \vect{h}_k(\vect{m}_{k \mid k-1}), \\
  \vect{S}_k &= \vect{H}_\vect{x}(\vect{m}_{k \mid k-1}) \, \vect{P}_{k \mid k-1} \, \vect{H}_\vect{x}\T(\vect{m}_{k \mid k-1}) + \vectb{\Sigma}_k, \\
  \vect{K}_k &= \vect{P}_{k \mid k-1} \, \vect{H}_\vect{x}\T(\vect{m}_{k \mid k-1}) \,  \vect{S}_k^{-1}, \\
  \vect{m}_{k \mid k} &= \vect{m}_{k \mid k-1} + \vect{K}_k \, \vect{v}_k, \\
  \vect{P}_{k \mid k} &= [\vect{I}-\vect{K}_k \, \vect{H}_\vect{x}(\vect{m}_{k \mid k-1})] \, 
\vect{P}_{k \mid k-1} \,  \\ & \qquad[\vect{I}-\vect{K}_k \, \vect{H}_\vect{x}(\vect{m}_{k \mid k-1})]\T + \vect{K}_k \, \vectb{\Sigma}_k \, \vect{K}_k\T,
\end{split}
\end{equation}
where $\vect{H}_\vect{x}(\cdot)$ denotes the Jacobian of the measurement model $ \vect{h}_k(\cdot)$ with respect to the state variables $\vect{x}$. The slightly unorthodox form of the last line is known as the Joseph's formula, which both numerically stabilizes updating the covariance and preserves symmetry.

The linearizations inside the extended Kalman filter cause some errors in the estimation. Most notably the estimation scheme does not preserve the norm of the orientation quaternions. Therefore after each update an extra quaternion normalization step is added to the estimation scheme.

\subsection{Propagation by Gyroscope Prediction}
\noindent
The state holds the internal and external parameters of the camera and the 3D position of the features. The internal parameters and the 3D coordinates of the features are assumed to stay constant (but are still unknown), so their propagation functions are identities. The external parameters (position, orientation) of the camera are not constant and are treated differently.

The position of the camera is modeled as a Wiener velocity process, a commonly used model in tracking and control literature (see, \eg, \cite{Sarkka:2013} for details). In order to keep track of the full inertial state the estimate contains both the position and velocity vectors $ \vect{x} = \begin{pmatrix} \vect{p} & \vect{v} \end{pmatrix}\T$ and the acceleration is modeled as a white noise process $\frac{\diff^2 x(t)}{\diff t^2} = w(t)$, or in state space form as a linear time-invariant stochastic differential equation (independently for each spatial dimension)
\begin{equation}
  \frac{\diff\vect{x}(t)}{\diff t} =
  \begin{pmatrix}
    0 & 1 \\
    0 & 0 \\
  \end{pmatrix} 
  \vect{x}(t) + 
  \begin{pmatrix}
    0 \\
    1 \\
  \end{pmatrix}   
  w(t),
\end{equation}
where $w(t)$ is a realization of the white noise process. In discrete time the system is  
\begin{equation}
  \vect{x}_{k+1} = \vect{A}\vect{x}_{k}+\vectb{\varepsilon}_k \label{eq:discrete}
\end{equation}
where $\vectb{\varepsilon}_k \sim \mathrm{N}(\vect{0},\vect{Q}_t)$, and $\vect{x}_k := \vect{x}(t_k)$.

The orientation of the camera is encoded in a unit quaternion, unit quaternions are a direct representation of an axis-angle rotation and can be converted into a rotation matrix using Rodriguez formula.

Given a unit quaternion $\vect{q}$ that represents a rotation and a known angular rate $\vectb{\omega}$ in the same frame of reference, its derivative can be expressed as 
\begin{equation}
  \frac{\diff\vect{q}(t)}{\diff t}=\frac{1}{2}\,\Omega(\vectb{\omega})\,\vect{q}(t),
\end{equation}
where 
\begin{equation}
  \Omega(\vectb{\omega}) =
  \begin{pmatrix}
    0 & -\vectb{\omega}\T  \\
    \vectb{\omega} & [\vectb{\omega}]_\times
  \end{pmatrix}.
\end{equation}
The notation $[w]_\times$ is the $3 \times 3$ cross-product matrix (for further details on quaternion modeling, see, \eg, \cite{Kok:2014}).

Assuming constant rotation rate (during $\Delta t$), the discrete-time system is 
\begin{equation}
  \vect{q}_{k+1}=\exp\bigg(\frac{\Delta t_k \, \Omega(\vectb{\omega}_k)}{2}\bigg)\,\vect{q}_{k}.
\end{equation}

Since the gyroscope produces rotational rate measurements with known accuracy, it can be propagated into an uncertainty for the quaternion. The gyroscope data is used directly in the prediction as a control signal in a linear Kalman filter.

Putting it all together, the state dynamics are described by
\begin{equation}
\vect{f}_k(\vect{x}_k, \vectb{\varepsilon}_k)=\vect{A}_k\vect{x}+\vectb{\varepsilon}_k,
\end{equation}
where the linear dynamics are given by
\begin{equation}
\vect{A}_k =\begin{pmatrix}
    \vect{I}_3 & &  & & \\
      & \vect{I}_3 & \vect{I}_3\Delta t_k & &  \\ 
      &  & \vect{I}_3 & & \\ 
      &  &  & \exp(\frac{\Delta t_k}{2} \, \Omega(\vectb{\omega}_k-\vectb{\omega}_b)) &  \\ 
      &  &  &  & \vect{I}_3
  \end{pmatrix}
\end{equation}
and 
  \begin{equation}
   \vectb{\varepsilon}_k \sim \mathrm{N}(\vect{0},\vect{Q}), \quad
 \vect{Q} =\operatorname{blkdiag}\begin{pmatrix}
    \vect{0}_3 &
    \vect{Q}_{t} &
    \vect{Q}_{q}& 
    \vect{0}_3
  \end{pmatrix},
\end{equation}
where $\vectb{\omega}_b$ is the gyroscope bias (estimated off-line) and the $\vect{Q}_{t}$ and $\vect{Q}_{q}$ matrices model the process noise of the translation and rotation, respectively.

The process noise of the translation is derived from the wiener velocity model described above, the process noise of the rotation is propagated from the rotational rate.

\subsection{Visual Update}
\noindent
The visual update couples all of the state variables. In Figure~\ref{fig:flow} the visual update occurs every time a new frame has been acquired. The frame is first processed by the feature tracker and inlier detection, and then the feature pixel coordinates are passed to the visual update model which processes an extended Kalman filter update step.

The features are chosen by the Shi--Tomasi {\em Good features to track} method \cite{Shi+Tomasi:1994} which determines strong corners in the image. These features are tracked across frames by a pyramidal {\em Lucas--Kanade tracker} \cite{Lucas+Kanade:1981,Bouguet:2001}. The {\em Seven-point algorithm} \cite{Hartley+Zisserman:2004} is used for inlier detection. If a feature is recognized as an inlier, the visual update directly proceeds for the feature. If the feature is not an inlier, the feature is replaced in the state by re-initializing its current position $\vect{z}^{(i)}$ estimate (both in terms of state mean and covariance) to uninformative \textit{a~priori} values.

The measurement model function $\vect{h}(\cdot)$ is a function of all state variables. The measurement vector $\vect{y} \in \R^{2d}$ contains the pixel coordinates, and feature-wise the model can be written as
\begin{equation}
  \vect{y}^{(i)} = \vect{h}^{(i)}(\vect{x}) = \mathrm{distort}\!\left[\vect{K}(\vect{c}) \, \vect{E}(\vect{p}, \vect{q}) \begin{pmatrix} \vect{z}^{(i)} \\ 1 \end{pmatrix}, \vect{c}\,\right],
  \label{eq:update}
\end{equation}
for $i=1,2,\ldots,d$, where $d$ is the number of features being tracked. The measurement model is fully determined by the camera model in Eqs.~\eqref{eq:pinhole1}--\eqref{eq:radial2}. For the EKF update, the Jacobian matrix of Eq.~\eqref{eq:update} must be formed as well. The measurement noise is set to $\vectb{\Sigma} = (2.5\,\text{px})^2 \, \vect{I}$.

\section{Experiments}
\noindent
We demonstrate our self-calibration method both in a simulation setup with known ground-truth values, and empirically on real data. 
All the experiments runs were performed on the data off-line using a Mathworks \textsc{Matlab} implementation\footnote{Code available online:\\ \mbox{\url{https://github.com/AaltoVision/camera-gyro-calibration}}}, and benchmarked against the publicly available \textsc{Matlab} implementation by Jia and Evans \cite{Jia+Evans:2014}. Furthermore, we compared our approach to a bundle-adjustment setup, which was used for checking how well the other methods actually were able to perform.

\begin{table}
  \caption{RMSE errors per method and parameter.}
  \label{tbl:results}
  \centering\footnotesize
  \begin{tabular}{ l c c c c c c c c } 
  \toprule
  Variable & Initial & Batch opt. & Jia and Evans \cite{Jia+Evans:2014} & Proposed \\
  \midrule
  $f_x$ (px) & 	75 & 	0.05 & 15.19	 & 	0.36\\
  $f_y$ (px) & 	75& 	0.05&  15.19	& 	0.38\\
  $c_x$ (px) & 	0.5 & 	0.02 & 	1.91 & 	0.27 \\
  $c_y$ (px) & 	0.5 & 	0.10& 	2.09& 	0.34 \\
  $k_1$  &     	0.01& 	0.0001 &	--- & 	0.0001 \\
  $k_2$  &     	0.01 & 	0.0095 & ---	&	0.0095 \\
  \bottomrule
  \end{tabular}
\end{table}

\subsection{Simulation Study}
\noindent
A simulation similar to the one described in \cite{Jia+Evans:2014} was performed to compare the results. To evaluate the methods, 100 camera movement paths around a three-dimensional regular point structure were simulated in a Monte Carlo setup. The point coordinates were projected into a virtual camera that followed the paths and continuously turned to look at the point structure. The ground-truth camera model was a zero-skew square pixel camera with a focal length of 575~pixels, an image size of $480 \times 640$ pixels, with the origin at the center of the frame. The distortion parameters were set to be zero in the simulation. 

The simulation setup produced tracks for the visual features and gyroscope readings, at 10~Hz and 100~Hz, respectively. The initial state estimates corresponding to the camera parameters were set as follows. The initial guess for the origin of the image plane and focal length were the image center coordinates and 700~px, respectively.

Table~\ref{tbl:results} shows the average RMSE results over the 100 simulations for the initial state and three methods run on the simulated data. The first column shows the initial RMSE from the initialized parameters. We compare the calibration parameters acquired from three models. The first is the model proposed, which was initialized and run as described earlier in the Methods section.

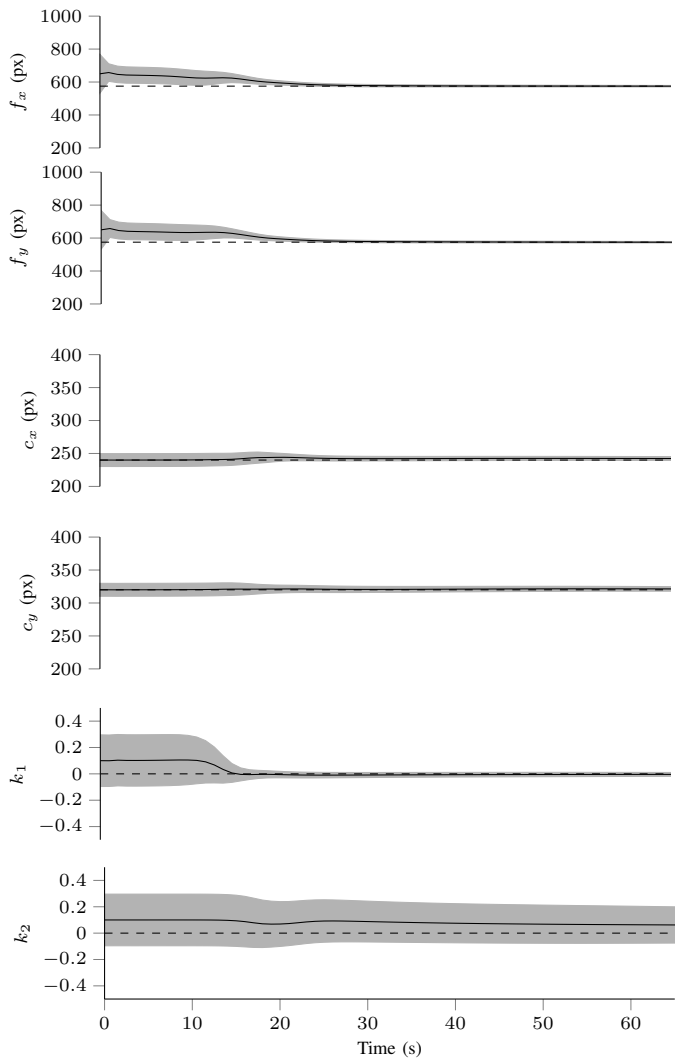
\begin{figure}[!t]

  \setlength{\figurewidth}{.90\columnwidth}
  \setlength{\figureheight}{0.22\figurewidth}
  \raggedleft\scriptsize%
  {\pgfplotsset{axis on top, hide x axis, ytick align=outside, xmin=0, xmax=65}
%
%
\begin{tikzpicture}

\begin{axis}[%
width=0.951\figurewidth,
height=\figureheight,
at={(0\figurewidth,0\figureheight)},
scale only axis,
xmin=0,
ymin=200,
ymax=1000,
ylabel={$f_x$ (px)},
axis background/.style={fill=white},
axis x line*=bottom,
axis y line*=left
]

\addplot[area legend,solid,draw=white!70!black,fill=white!70!black,forget plot]
table[row sep=crcr] {%
x	y\\
0	767.6\\
1	710.78481555832\\
2	696.493394852538\\
3	692.084993611949\\
4	690.76793926652\\
5	689.476136057243\\
6	688.065674516487\\
7	686.082903204457\\
8	683.212017623312\\
9	679.192418744626\\
10	674.203146219101\\
11	669.3483323371\\
12	665.833360272858\\
13	662.93993666089\\
14	658.84249376297\\
15	651.825265505599\\
16	642.307942401183\\
17	631.919137065053\\
18	622.38524848194\\
19	615.275244431611\\
20	610.018036138648\\
21	605.461823263789\\
22	601.280753387526\\
23	597.626895705222\\
24	594.580739951862\\
25	592.135007133892\\
26	590.185969152766\\
27	588.639102954651\\
28	587.4090384677\\
29	586.419205781712\\
30	585.61564595536\\
31	584.957550651018\\
32	584.410996956857\\
33	583.949371476933\\
34	583.550997779044\\
35	583.203093003534\\
36	582.895997082211\\
37	582.623766077432\\
38	582.3816017135\\
39	582.165210088754\\
40	581.970230889153\\
41	581.792966883042\\
42	581.631275321609\\
43	581.482760934028\\
44	581.345260164353\\
45	581.217453846779\\
46	581.097698381915\\
47	580.984865082919\\
48	580.877558836926\\
49	580.774783742326\\
50	580.675668281649\\
51	580.579384361054\\
52	580.485329006903\\
53	580.392116096038\\
54	580.299613056246\\
55	580.206654216675\\
56	580.113479730499\\
57	580.019628183974\\
58	579.925442711253\\
59	579.830501289695\\
60	579.734312412968\\
61	579.636468820137\\
62	579.536677728822\\
63	579.435463864755\\
64	579.333006415051\\
65	579.229131139198\\
66	579.123504681701\\
67	579.015792279112\\
68	578.906356325726\\
69	578.794638601049\\
70	578.681393046031\\
71	578.565852277086\\
72	578.448486976389\\
73	578.328459369255\\
74	578.205073773762\\
75	578.079561987473\\
76	577.952194778331\\
77	577.82284032852\\
78	577.691492462662\\
79	577.557135361727\\
80	577.419636313764\\
81	577.27869368935\\
82	577.133298222361\\
83	576.979397210592\\
84	576.814609314673\\
85	576.644124215766\\
86	576.476169530908\\
87	576.315793128427\\
88	576.165005892452\\
89	576.028220570792\\
90	575.912131129864\\
91	575.825678562576\\
92	575.773893871706\\
93	575.765655676637\\
94	575.808290511213\\
95	575.892317131484\\
96	575.996632458375\\
97	576.09490991066\\
98	576.171838722099\\
99	576.225262244869\\
100	576.255278706359\\
100	573.850376986047\\
99	573.771981524312\\
98	573.638726002115\\
97	573.456755383732\\
96	573.220023955777\\
95	572.935022846584\\
94	572.633724058046\\
93	572.348707929968\\
92	572.115275182911\\
91	571.930206506107\\
90	571.776144550119\\
89	571.636701837822\\
88	571.506299112024\\
87	571.383536823439\\
86	571.267617074404\\
85	571.15825214229\\
84	571.056879199942\\
83	570.966595642486\\
82	570.890031148382\\
81	570.826613794881\\
80	570.774309830487\\
79	570.73162005327\\
78	570.696593990066\\
77	570.667793189434\\
76	570.64363339835\\
75	570.623223516077\\
74	570.605806249904\\
73	570.591005733632\\
72	570.577970662103\\
71	570.565776478289\\
70	570.553859127004\\
69	570.542027085817\\
68	570.529716370811\\
67	570.517194459409\\
66	570.504448177168\\
65	570.491550454199\\
64	570.478264099686\\
63	570.464768441932\\
62	570.451092869054\\
61	570.437824501213\\
60	570.425023444233\\
59	570.412660072767\\
58	570.400930674058\\
57	570.389557794207\\
56	570.378783561507\\
55	570.368602474729\\
54	570.359570850048\\
53	570.350948124648\\
52	570.343520216344\\
51	570.336723198278\\
50	570.330674653884\\
49	570.325625517349\\
48	570.321376970609\\
47	570.317999029338\\
46	570.315742056784\\
45	570.314933377398\\
44	570.315598729988\\
43	570.318690271288\\
42	570.324533620066\\
41	570.33374096247\\
40	570.347022381435\\
39	570.36605750774\\
38	570.391843559216\\
37	570.426126425922\\
36	570.471754944494\\
35	570.532132740496\\
34	570.610548669288\\
33	570.711411366047\\
32	570.839465450532\\
31	571.003181811999\\
30	571.215488868809\\
29	571.495309330481\\
28	571.866084027352\\
27	572.357078078963\\
26	573.014330765697\\
25	573.893964714815\\
24	575.059775508412\\
23	576.581360970626\\
22	578.456804181547\\
21	580.570297558109\\
20	582.61026460184\\
19	584.47557980648\\
18	586.919009907076\\
17	590.636626187083\\
16	594.661311407516\\
15	596.300496929724\\
14	592.849358693364\\
13	586.287250069098\\
12	581.434520253614\\
11	580.093348596212\\
10	581.750274200032\\
9	584.636800814382\\
8	587.332689941585\\
7	589.355093315413\\
6	590.729323455405\\
5	591.625455078543\\
4	592.350577640368\\
3	592.957447441407\\
2	596.114407712731\\
1	604.301360637359\\
0	532.4\\
}--cycle;
\addplot [color=black,solid,forget plot]
  table[row sep=crcr]{%
0	650\\
1	657.543088097839\\
2	646.303901282635\\
3	642.521220526678\\
4	641.559258453444\\
5	640.550795567893\\
6	639.397498985946\\
7	637.718998259935\\
8	635.272353782449\\
9	631.914609779504\\
10	627.976710209567\\
11	624.720840466656\\
12	623.633940263236\\
13	624.613593364994\\
14	625.845926228167\\
15	624.062881217661\\
16	618.48462690435\\
17	611.277881626068\\
18	604.652129194508\\
19	599.875412119045\\
20	596.314150370244\\
21	593.016060410949\\
22	589.868778784536\\
23	587.104128337924\\
24	584.820257730137\\
25	583.014485924354\\
26	581.600149959231\\
27	580.498090516807\\
28	579.637561247526\\
29	578.957257556096\\
30	578.415567412084\\
31	577.980366231509\\
32	577.625231203695\\
33	577.33039142149\\
34	577.080773224166\\
35	576.867612872015\\
36	576.683876013353\\
37	576.524946251677\\
38	576.386722636358\\
39	576.265633798247\\
40	576.158626635294\\
41	576.063353922756\\
42	575.977904470837\\
43	575.900725602658\\
44	575.83042944717\\
45	575.766193612088\\
46	575.706720219349\\
47	575.651432056129\\
48	575.599467903768\\
49	575.550204629838\\
50	575.503171467766\\
51	575.458053779666\\
52	575.414424611623\\
53	575.371532110343\\
54	575.329591953147\\
55	575.287628345702\\
56	575.246131646003\\
57	575.204592989091\\
58	575.163186692656\\
59	575.121580681231\\
60	575.0796679286\\
61	575.037146660675\\
62	574.993885298938\\
63	574.950116153344\\
64	574.905635257369\\
65	574.860340796698\\
66	574.813976429434\\
67	574.766493369261\\
68	574.718036348268\\
69	574.668332843433\\
70	574.617626086517\\
71	574.565814377688\\
72	574.513228819246\\
73	574.459732551444\\
74	574.405440011833\\
75	574.351392751775\\
76	574.297914088341\\
77	574.245316758977\\
78	574.194043226364\\
79	574.144377707499\\
80	574.096973072126\\
81	574.052653742116\\
82	574.011664685372\\
83	573.972996426539\\
84	573.935744257307\\
85	573.901188179028\\
86	573.871893302656\\
87	573.849664975933\\
88	573.835652502238\\
89	573.832461204307\\
90	573.844137839992\\
91	573.877942534341\\
92	573.944584527308\\
93	574.057181803302\\
94	574.221007284629\\
95	574.413669989034\\
96	574.608328207076\\
97	574.775832647196\\
98	574.905282362107\\
99	574.99862188459\\
100	575.052827846203\\
};
\addplot [color=black,dashed,forget plot]
  table[row sep=crcr]{%
0	575\\
100	575\\
};
\end{axis}
\end{tikzpicture}%
  \vspace*{10pt}
%
%
\begin{tikzpicture}

\begin{axis}[%
width=0.951\figurewidth,
height=\figureheight,
at={(0\figurewidth,0\figureheight)},
scale only axis,
xmin=0,
ymin=200,
ymax=1000,
ylabel={$f_y$ (px)},
axis background/.style={fill=white},
axis x line*=bottom,
axis y line*=left
]

\addplot[area legend,solid,draw=white!70!black,fill=white!70!black,forget plot]
table[row sep=crcr] {%
x	y\\
0	767.6\\
1	712.04675709561\\
2	696.786489842613\\
3	691.529098634935\\
4	689.874578095016\\
5	688.619907328088\\
6	687.382804795078\\
7	685.748524668084\\
8	683.77883263717\\
9	681.820894248658\\
10	680.255830480503\\
11	678.722539655512\\
12	676.744763353119\\
13	672.794596668577\\
14	665.45239295432\\
15	655.60953745836\\
16	644.686785870629\\
17	633.5479142621\\
18	623.245543673099\\
19	615.207333858091\\
20	609.62203132874\\
21	605.162704551886\\
22	601.153384007683\\
23	597.654907038556\\
24	594.742190774271\\
25	592.407496043223\\
26	590.548394906442\\
27	589.070781250931\\
28	587.891750652746\\
29	586.937395039182\\
30	586.156445933116\\
31	585.510514564588\\
32	584.968365281059\\
33	584.505048223126\\
34	584.099884920964\\
35	583.740603477689\\
36	583.417937268143\\
37	583.126520359937\\
38	582.862517594326\\
39	582.622574009886\\
40	582.402769104957\\
41	582.199492919628\\
42	582.011224924693\\
43	581.835834583277\\
44	581.67127655692\\
45	581.516387698485\\
46	581.36968890166\\
47	581.230157957974\\
48	581.096393114034\\
49	580.967571562891\\
50	580.842765561829\\
51	580.721088269289\\
52	580.601971614152\\
53	580.483876674382\\
54	580.36661312299\\
55	580.248975853903\\
56	580.131182746016\\
57	580.012849457017\\
58	579.894484902484\\
59	579.775746376533\\
60	579.656184088458\\
61	579.535444442433\\
62	579.413242967396\\
63	579.290198521549\\
64	579.166655708449\\
65	579.042592285393\\
66	578.917791343478\\
67	578.791941489955\\
68	578.665433586454\\
69	578.537754659369\\
70	578.409807711555\\
71	578.280781321474\\
72	578.15113643483\\
73	578.019965262356\\
74	577.88642874976\\
75	577.751757612477\\
76	577.616327269583\\
77	577.48010464602\\
78	577.343177164981\\
79	577.204660210334\\
80	577.064512199967\\
81	576.922451222687\\
82	576.777591710862\\
83	576.626074377215\\
84	576.465910523243\\
85	576.302662719039\\
86	576.144860811653\\
87	575.997873066456\\
88	575.864041196764\\
89	575.74752907417\\
90	575.653397600733\\
91	575.587924682281\\
92	575.553736393541\\
93	575.557106983812\\
94	575.604135564484\\
95	575.68643188914\\
96	575.785325247905\\
97	575.876414575986\\
98	575.946982140119\\
99	575.996213698221\\
100	576.024752249003\\
100	573.536564004594\\
99	573.461651662801\\
98	573.338745160563\\
97	573.176040381926\\
96	572.969570512438\\
95	572.72362324801\\
94	572.464013919655\\
93	572.216930133398\\
92	572.010046025316\\
91	571.837638062078\\
90	571.683851792526\\
89	571.535158869795\\
88	571.388802265734\\
87	571.245005442654\\
86	571.104707666722\\
85	570.969783890454\\
84	570.843769767492\\
83	570.731607295476\\
82	570.636883075889\\
81	570.558731560083\\
80	570.494555150346\\
79	570.44265077883\\
78	570.400991686888\\
77	570.367962322877\\
76	570.341682515426\\
75	570.321132141318\\
74	570.305368482868\\
73	570.293910674628\\
72	570.285924108097\\
71	570.280527874848\\
70	570.27717998615\\
69	570.27570966927\\
68	570.27559535401\\
67	570.277018687208\\
66	570.279881786201\\
65	570.284264147191\\
64	570.289913484315\\
63	570.296914150941\\
62	570.305137755505\\
61	570.314935906798\\
60	570.326228365008\\
59	570.338947891023\\
58	570.353198482646\\
57	570.368642474327\\
56	570.385406535216\\
55	570.403271997727\\
54	570.422612869503\\
53	570.44268272995\\
52	570.464278107518\\
51	570.486893252418\\
50	570.510841571454\\
49	570.536311873089\\
48	570.563198810896\\
47	570.591681855676\\
46	570.621828133228\\
45	570.6539556468\\
44	570.688052760803\\
43	570.724858149309\\
42	570.764551470352\\
41	570.807668231584\\
40	570.854663440225\\
39	570.906513287994\\
38	570.964163007731\\
37	571.028985004047\\
36	571.102778746064\\
35	571.187874900522\\
34	571.286824291175\\
33	571.40365589632\\
32	571.542722879999\\
31	571.711721158232\\
30	571.922091701986\\
29	572.190134581229\\
28	572.53672186054\\
27	572.98830600212\\
26	573.588189425218\\
25	574.39033086233\\
24	575.458392732445\\
23	576.862105113059\\
22	578.603529386616\\
21	580.578989329955\\
20	582.553149717823\\
19	584.645445326125\\
18	587.659217151356\\
17	591.851343189995\\
16	596.652133015112\\
15	600.415958909888\\
14	600.71231490843\\
13	597.448951405254\\
12	592.96095416369\\
11	589.232256562913\\
10	586.782004496962\\
9	585.7191064221\\
8	586.150777237895\\
7	587.197005626015\\
6	588.201693976489\\
5	588.913811044244\\
4	589.593667064898\\
3	590.504036319315\\
2	594.462364721148\\
1	603.537853929107\\
0	532.4\\
}--cycle;
\addplot [color=black,solid,forget plot]
  table[row sep=crcr]{%
0	650\\
1	657.792305512358\\
2	645.62442728188\\
3	641.016567477125\\
4	639.734122579957\\
5	638.766859186166\\
6	637.792249385784\\
7	636.472765147049\\
8	634.964804937532\\
9	633.770000335379\\
10	633.518917488733\\
11	633.977398109213\\
12	634.852858758404\\
13	635.121774036916\\
14	633.082353931375\\
15	628.012748184124\\
16	620.66945944287\\
17	612.699628726048\\
18	605.452380412227\\
19	599.926389592108\\
20	596.087590523282\\
21	592.870846940921\\
22	589.87845669715\\
23	587.258506075807\\
24	585.100291753358\\
25	583.398913452776\\
26	582.06829216583\\
27	581.029543626526\\
28	580.214236256643\\
29	579.563764810206\\
30	579.039268817551\\
31	578.61111786141\\
32	578.255544080529\\
33	577.954352059723\\
34	577.693354606069\\
35	577.464239189106\\
36	577.260358007104\\
37	577.077752681992\\
38	576.913340301029\\
39	576.76454364894\\
40	576.628716272591\\
41	576.503580575606\\
42	576.387888197522\\
43	576.280346366293\\
44	576.179664658861\\
45	576.085171672643\\
46	575.995758517444\\
47	575.910919906825\\
48	575.829795962465\\
49	575.75194171799\\
50	575.676803566641\\
51	575.603990760854\\
52	575.533124860835\\
53	575.463279702166\\
54	575.394612996246\\
55	575.326123925815\\
56	575.258294640616\\
57	575.190745965672\\
58	575.123841692565\\
59	575.057347133778\\
60	574.991206226733\\
61	574.925190174616\\
62	574.859190361451\\
63	574.793556336245\\
64	574.728284596382\\
65	574.663428216292\\
66	574.598836564839\\
67	574.534480088582\\
68	574.470514470232\\
69	574.406732164319\\
70	574.343493848853\\
71	574.280654598161\\
72	574.218530271463\\
73	574.156937968492\\
74	574.095898616314\\
75	574.036444876898\\
76	573.979004892504\\
77	573.924033484449\\
78	573.872084425935\\
79	573.823655494582\\
80	573.779533675157\\
81	573.740591391385\\
82	573.707237393375\\
83	573.678840836345\\
84	573.654840145368\\
85	573.636223304747\\
86	573.624784239187\\
87	573.621439254555\\
88	573.626421731249\\
89	573.641343971983\\
90	573.66862469663\\
91	573.712781372179\\
92	573.781891209429\\
93	573.887018558605\\
94	574.034074742069\\
95	574.205027568575\\
96	574.377447880172\\
97	574.526227478956\\
98	574.642863650341\\
99	574.728932680511\\
100	574.780658126798\\
};
\addplot [color=black,dashed,forget plot]
  table[row sep=crcr]{%
0	575\\
100	575\\
};
\end{axis}
\end{tikzpicture}%
  \vspace*{10pt}
%
%
\begin{tikzpicture}

\begin{axis}[%
width=0.951\figurewidth,
height=\figureheight,
at={(0\figurewidth,0\figureheight)},
scale only axis,
xmin=0,
ymin=200,
ymax=400,
ylabel={$c_x$ (px)},
axis background/.style={fill=white},
axis x line*=bottom,
axis y line*=left
]

\addplot[area legend,solid,draw=white!70!black,fill=white!70!black,forget plot]
table[row sep=crcr] {%
x	y\\
0	249.8\\
1	249.789908922602\\
2	249.747016498489\\
3	249.808676623582\\
4	249.818401549662\\
5	249.817073707887\\
6	249.820256200371\\
7	249.830034107224\\
8	249.847902728924\\
9	249.871977615636\\
10	249.908236426879\\
11	249.978411373727\\
12	250.111127534913\\
13	250.252137225035\\
14	250.252397817631\\
15	250.397246134359\\
16	251.041879666643\\
17	251.853658351271\\
18	252.192596869913\\
19	251.485523537557\\
20	250.645860072916\\
21	249.92480913514\\
22	249.052524143649\\
23	248.171117814479\\
24	247.422071645522\\
25	246.84667638224\\
26	246.427482912883\\
27	246.131068700816\\
28	245.92505733249\\
29	245.782302020587\\
30	245.685688108811\\
31	245.622300565305\\
32	245.580830197387\\
33	245.554889564165\\
34	245.539906915857\\
35	245.533340293111\\
36	245.533385252273\\
37	245.537885872067\\
38	245.544646635422\\
39	245.552109561505\\
40	245.559864024356\\
41	245.567378336374\\
42	245.573710079399\\
43	245.578650191155\\
44	245.582201659527\\
45	245.584138377359\\
46	245.584377387763\\
47	245.583000038095\\
48	245.579808022312\\
49	245.574747944719\\
50	245.568257477296\\
51	245.560129845473\\
52	245.550326352355\\
53	245.538928903437\\
54	245.525837819495\\
55	245.510912093696\\
56	245.493792216933\\
57	245.474608638229\\
58	245.453516690219\\
59	245.430606986945\\
60	245.405934378096\\
61	245.379607284374\\
62	245.351544265098\\
63	245.322022737722\\
64	245.291155347024\\
65	245.259188896238\\
66	245.226328321708\\
67	245.192487104527\\
68	245.157767082609\\
69	245.122056999664\\
70	245.085923558803\\
71	245.04886223317\\
72	245.010843253062\\
73	244.971048434351\\
74	244.927972978463\\
75	244.880573094933\\
76	244.828422072491\\
77	244.770906625513\\
78	244.706922018234\\
79	244.634988518146\\
80	244.553808317074\\
81	244.460502499711\\
82	244.354115495222\\
83	244.231901104051\\
84	244.095793992632\\
85	243.953142728178\\
86	243.810432671172\\
87	243.670110947878\\
88	243.534323579556\\
89	243.407740185998\\
90	243.297907304334\\
91	243.211914207833\\
92	243.143372228246\\
93	243.076870698441\\
94	242.98641781194\\
95	242.85475284757\\
96	242.680352455687\\
97	242.493452809615\\
98	242.326528890608\\
99	242.19271581846\\
100	242.108055308847\\
100	239.640121040837\\
99	239.671592815712\\
98	239.713371117676\\
97	239.753432187184\\
96	239.781184688504\\
95	239.781664745146\\
94	239.741265173622\\
93	239.658559256156\\
92	239.550063688513\\
91	239.438851156549\\
90	239.337516206106\\
89	239.250438991209\\
88	239.18032631796\\
87	239.130387701669\\
86	239.100570852859\\
85	239.087879961794\\
84	239.089916049347\\
83	239.103227736992\\
82	239.1222654695\\
81	239.142308056689\\
80	239.161754537015\\
79	239.179460388753\\
78	239.195865709532\\
77	239.211054187091\\
76	239.225092268604\\
75	239.238230909405\\
74	239.250488964521\\
73	239.26182521532\\
72	239.27282831466\\
71	239.284060232511\\
70	239.295702900636\\
69	239.307710817493\\
68	239.320119760077\\
67	239.332552818972\\
66	239.344926923954\\
65	239.357125799836\\
64	239.369016611935\\
63	239.380312340882\\
62	239.390673233963\\
61	239.400038147401\\
60	239.408165575204\\
59	239.414996018174\\
58	239.420323250103\\
57	239.423903322467\\
56	239.425556826975\\
55	239.425145777039\\
54	239.422449251078\\
53	239.417775638341\\
52	239.411203447607\\
51	239.40271612137\\
50	239.391980622832\\
49	239.378962613854\\
48	239.363695057513\\
47	239.345526393195\\
46	239.324345318984\\
45	239.300088657764\\
44	239.272475689705\\
43	239.241338480971\\
42	239.206748172459\\
41	239.168238278083\\
40	239.125656644884\\
39	239.079848084277\\
38	239.030668763585\\
37	238.977834847563\\
36	238.92244526473\\
35	238.866358002516\\
34	238.811170846013\\
33	238.757619510375\\
32	238.706312447665\\
31	238.65845891143\\
30	238.616091012127\\
29	238.584740911823\\
28	238.569929795328\\
27	238.576965865828\\
26	238.614914190859\\
25	238.686292642648\\
24	238.783346135411\\
23	238.864015043745\\
22	238.818471397793\\
21	238.434347167596\\
20	237.505419967207\\
19	236.220641066066\\
18	235.186086351512\\
17	234.078916483771\\
16	232.860632834545\\
15	231.879734588328\\
14	231.392179230721\\
13	231.131381245978\\
12	230.828372391467\\
11	230.578332966716\\
10	230.427733160858\\
9	230.343303025207\\
8	230.292478391717\\
7	230.259864984143\\
6	230.241990887113\\
5	230.234472605155\\
4	230.233268702403\\
3	230.22124944454\\
2	230.156192612229\\
1	230.19412930296\\
0	230.2\\
}--cycle;
\addplot [color=black,solid,forget plot]
  table[row sep=crcr]{%
0	240\\
1	239.992019112781\\
2	239.951604555359\\
3	240.014963034061\\
4	240.025835126033\\
5	240.025773156521\\
6	240.031123543742\\
7	240.044949545683\\
8	240.070190560321\\
9	240.107640320422\\
10	240.167984793868\\
11	240.278372170222\\
12	240.46974996319\\
13	240.691759235507\\
14	240.822288524176\\
15	241.138490361343\\
16	241.951256250594\\
17	242.966287417521\\
18	243.689341610713\\
19	243.853082301811\\
20	244.075640020062\\
21	244.179578151368\\
22	243.935497770721\\
23	243.517566429112\\
24	243.102708890466\\
25	242.766484512444\\
26	242.521198551871\\
27	242.354017283322\\
28	242.247493563909\\
29	242.183521466205\\
30	242.150889560469\\
31	242.140379738368\\
32	242.143571322526\\
33	242.15625453727\\
34	242.175538880935\\
35	242.199849147814\\
36	242.227915258501\\
37	242.257860359815\\
38	242.287657699503\\
39	242.315978822891\\
40	242.34276033462\\
41	242.367808307229\\
42	242.390229125929\\
43	242.409994336063\\
44	242.427338674616\\
45	242.442113517562\\
46	242.454361353373\\
47	242.464263215645\\
48	242.471751539913\\
49	242.476855279287\\
50	242.480119050064\\
51	242.481422983421\\
52	242.480764899981\\
53	242.478352270889\\
54	242.474143535287\\
55	242.468028935368\\
56	242.459674521954\\
57	242.449255980348\\
58	242.436919970161\\
59	242.42280150256\\
60	242.40704997665\\
61	242.389822715887\\
62	242.371108749531\\
63	242.351167539302\\
64	242.33008597948\\
65	242.308157348037\\
66	242.285627622831\\
67	242.262519961749\\
68	242.238943421343\\
69	242.214883908579\\
70	242.190813229719\\
71	242.16646123284\\
72	242.141835783861\\
73	242.116436824835\\
74	242.089230971492\\
75	242.059402002169\\
76	242.026757170548\\
77	241.990980406302\\
78	241.951393863883\\
79	241.90722445345\\
80	241.857781427044\\
81	241.8014052782\\
82	241.738190482361\\
83	241.667564420521\\
84	241.59285502099\\
85	241.520511344986\\
86	241.455501762016\\
87	241.400249324773\\
88	241.357324948758\\
89	241.329089588603\\
90	241.31771175522\\
91	241.325382682191\\
92	241.34671795838\\
93	241.367714977298\\
94	241.363841492781\\
95	241.318208796358\\
96	241.230768572096\\
97	241.1234424984\\
98	241.019950004142\\
99	240.932154317086\\
100	240.874088174842\\
};
\addplot [color=black,dashed,forget plot]
  table[row sep=crcr]{%
0	240\\
100	240\\
};
\end{axis}
\end{tikzpicture}%
  \vspace*{10pt}
%
%
\begin{tikzpicture}

\begin{axis}[%
width=0.951\figurewidth,
height=\figureheight,
at={(0\figurewidth,0\figureheight)},
scale only axis,
xmin=0,
ymin=200,
ymax=400,
ylabel={$c_y$ (px)},
axis background/.style={fill=white},
axis x line*=bottom,
axis y line*=left
]

\addplot[area legend,solid,draw=white!70!black,fill=white!70!black,forget plot]
table[row sep=crcr] {%
x	y\\
0	329.8\\
1	329.791341893723\\
2	329.741718498633\\
3	329.806248433874\\
4	329.822476070103\\
5	329.82825702488\\
6	329.84207218727\\
7	329.869365085537\\
8	329.912905241017\\
9	329.968296934799\\
10	330.030082480543\\
11	330.095425547554\\
12	330.180736157858\\
13	330.345870326894\\
14	330.551531736678\\
15	330.562071737754\\
16	330.156710464298\\
17	329.330865084676\\
18	328.383791154366\\
19	327.691926992314\\
20	327.249668311508\\
21	327.054515871506\\
22	326.927892234646\\
23	326.727683752752\\
24	326.462182873639\\
25	326.181192131995\\
26	325.921349025281\\
27	325.701794034496\\
28	325.527758554001\\
29	325.394421359288\\
30	325.296421682922\\
31	325.227854474414\\
32	325.180761009995\\
33	325.150095300918\\
34	325.132499560303\\
35	325.126053383422\\
36	325.129011553498\\
37	325.139920082629\\
38	325.156006812\\
39	325.175019954636\\
40	325.195572314782\\
41	325.217290482614\\
42	325.238593960938\\
43	325.259025272838\\
44	325.278006771528\\
45	325.2951735226\\
46	325.310223933982\\
47	325.322608698825\\
48	325.332288966184\\
49	325.339166558153\\
50	325.342956052977\\
51	325.343308639592\\
52	325.340068966199\\
53	325.332524175244\\
54	325.320254745583\\
55	325.302172937563\\
56	325.278243827991\\
57	325.248172319532\\
58	325.212101226758\\
59	325.169900958219\\
60	325.121392629027\\
61	325.066240505745\\
62	325.004531904524\\
63	324.93676523897\\
64	324.864040188061\\
65	324.787068016544\\
66	324.706358924388\\
67	324.622186878129\\
68	324.535507876843\\
69	324.446548592049\\
70	324.356723033771\\
71	324.265809483785\\
72	324.174520090877\\
73	324.082435726714\\
74	323.988730625883\\
75	323.894158529173\\
76	323.799291641083\\
77	323.704388892066\\
78	323.609822430126\\
79	323.515453444134\\
80	323.421445468242\\
81	323.327609109096\\
82	323.234349638303\\
83	323.139720461338\\
84	323.043439973876\\
85	322.949031775853\\
86	322.859603824699\\
87	322.773824601519\\
88	322.687496084148\\
89	322.595400637503\\
90	322.497483308559\\
91	322.399791386844\\
92	322.307745207793\\
93	322.227131546253\\
94	322.164259919724\\
95	322.116095926305\\
96	322.072505132372\\
97	322.029230503013\\
98	321.987392647174\\
99	321.950583866854\\
100	321.924668115443\\
100	319.015806687721\\
99	319.022567307204\\
98	319.025709588646\\
97	319.017823901823\\
96	318.989879604691\\
95	318.938148122272\\
94	318.868105549764\\
93	318.786662996036\\
92	318.707904024195\\
91	318.635571491074\\
90	318.56898481485\\
89	318.503661043462\\
88	318.435892963655\\
87	318.364577929491\\
86	318.291650015729\\
85	318.219870279493\\
84	318.150805930212\\
83	318.086556807083\\
82	318.029408097554\\
81	317.980441677989\\
80	317.939573277111\\
79	317.906647567759\\
78	317.88075768855\\
77	317.861265531403\\
76	317.847266769862\\
75	317.838109641866\\
74	317.833272087546\\
73	317.832324210876\\
72	317.834381740275\\
71	317.838584677656\\
70	317.844237662622\\
69	317.850786470635\\
68	317.857539332698\\
67	317.864064569635\\
66	317.869880619943\\
65	317.874289695904\\
64	317.876651403597\\
63	317.876519103707\\
62	317.873317588499\\
61	317.866483916166\\
60	317.855747036338\\
59	317.840680024872\\
58	317.821392095345\\
57	317.797615874412\\
56	317.769341527036\\
55	317.736568729216\\
54	317.699584463752\\
53	317.658008366355\\
52	317.612586533792\\
51	317.563310436304\\
50	317.510261066047\\
49	317.45343380876\\
48	317.392688272028\\
47	317.328011170736\\
46	317.259323508063\\
45	317.18628686206\\
44	317.109098514804\\
43	317.027946847216\\
42	316.942916931368\\
41	316.854170380818\\
40	316.761787736702\\
39	316.667477975289\\
38	316.570859188067\\
37	316.472791206754\\
36	316.375627243003\\
35	316.282199210082\\
34	316.193643736896\\
33	316.110589652445\\
32	316.033124540204\\
31	315.961712899798\\
30	315.898877981168\\
29	315.85039335257\\
28	315.820167540305\\
27	315.811117214247\\
26	315.825051739514\\
25	315.854033299039\\
24	315.876859748977\\
23	315.851887974491\\
22	315.719193140128\\
21	315.428597021159\\
20	315.042143963119\\
19	314.570778988141\\
18	313.655919345551\\
17	312.690628115227\\
16	311.961230175906\\
15	311.49000131804\\
14	311.15481476246\\
13	310.85897491464\\
12	310.664078547656\\
11	310.557369111025\\
10	310.473726215469\\
9	310.398797502687\\
8	310.335140683044\\
7	310.286642117765\\
6	310.256422272101\\
5	310.240933240071\\
4	310.233981608388\\
3	310.216247876254\\
2	310.149058629373\\
1	310.194806122504\\
0	310.2\\
}--cycle;
\addplot [color=black,solid,forget plot]
  table[row sep=crcr]{%
0	320\\
1	319.993074008113\\
2	319.945388564003\\
3	320.011248155064\\
4	320.028228839245\\
5	320.034595132475\\
6	320.049247229685\\
7	320.078003601651\\
8	320.12402296203\\
9	320.183547218743\\
10	320.251904348006\\
11	320.326397329289\\
12	320.422407352757\\
13	320.602422620767\\
14	320.853173249569\\
15	321.026036527897\\
16	321.058970320102\\
17	321.010746599952\\
18	321.019855249958\\
19	321.131352990228\\
20	321.145906137313\\
21	321.241556446333\\
22	321.323542687387\\
23	321.289785863622\\
24	321.169521311308\\
25	321.017612715517\\
26	320.873200382397\\
27	320.756455624371\\
28	320.673963047153\\
29	320.622407355929\\
30	320.597649832045\\
31	320.594783687106\\
32	320.606942775099\\
33	320.630342476682\\
34	320.6630716486\\
35	320.704126296752\\
36	320.75231939825\\
37	320.806355644691\\
38	320.863433000034\\
39	320.921248964963\\
40	320.978680025742\\
41	321.035730431716\\
42	321.090755446153\\
43	321.143486060027\\
44	321.193552643166\\
45	321.24073019233\\
46	321.284773721022\\
47	321.32530993478\\
48	321.362488619106\\
49	321.396300183456\\
50	321.426608559512\\
51	321.453309537948\\
52	321.476327749995\\
53	321.4952662708\\
54	321.509919604668\\
55	321.51937083339\\
56	321.523792677514\\
57	321.522894096972\\
58	321.516746661052\\
59	321.505290491545\\
60	321.488569832682\\
61	321.466362210956\\
62	321.438924746512\\
63	321.406642171338\\
64	321.370345795829\\
65	321.330678856224\\
66	321.288119772166\\
67	321.243125723882\\
68	321.19652360477\\
69	321.148667531342\\
70	321.100480348196\\
71	321.052197080721\\
72	321.004450915576\\
73	320.957379968795\\
74	320.911001356714\\
75	320.86613408552\\
76	320.823279205472\\
77	320.782827211734\\
78	320.745290059338\\
79	320.711050505947\\
80	320.680509372676\\
81	320.654025393543\\
82	320.631878867929\\
83	320.613138634211\\
84	320.597122952044\\
85	320.584451027673\\
86	320.575626920214\\
87	320.569201265505\\
88	320.561694523901\\
89	320.549530840482\\
90	320.533234061705\\
91	320.517681438959\\
92	320.507824615994\\
93	320.506897271145\\
94	320.516182734744\\
95	320.527122024289\\
96	320.531192368532\\
97	320.523527202418\\
98	320.50655111791\\
99	320.486575587029\\
100	320.470237401582\\
};
\addplot [color=black,dashed,forget plot]
  table[row sep=crcr]{%
0	320\\
100	320\\
};
\end{axis}
\end{tikzpicture}%
  \vspace*{10pt}
%
%
\begin{tikzpicture}

\begin{axis}[%
width=0.951\figurewidth,
height=\figureheight,
at={(0\figurewidth,0\figureheight)},
scale only axis,
xmin=0,
ymin=-0.5,
ymax=0.5,
ylabel={$k_1$},
axis background/.style={fill=white},
axis x line*=bottom,
axis y line*=left
]

\addplot[area legend,solid,draw=white!70!black,fill=white!70!black,forget plot]
table[row sep=crcr] {%
x	y\\
0	0.296\\
1	0.294668386518322\\
2	0.298678569300638\\
3	0.296688574773249\\
4	0.296449942934003\\
5	0.296666344281739\\
6	0.297151235639784\\
7	0.297695838780448\\
8	0.29792074688432\\
9	0.2969006384483\\
10	0.291965784840466\\
11	0.279344272460183\\
12	0.251946660978188\\
13	0.203685018883518\\
14	0.137214365237425\\
15	0.0785732114634312\\
16	0.0453386729429715\\
17	0.0313779901467954\\
18	0.0268105977834506\\
19	0.0240622148567131\\
20	0.0205303291775453\\
21	0.0173957078357077\\
22	0.0150309480114702\\
23	0.0133313382848577\\
24	0.0121964232682017\\
25	0.0114754852629301\\
26	0.0110351375431402\\
27	0.0107689743566538\\
28	0.0106198033236052\\
29	0.0105424812294945\\
30	0.010510064524261\\
31	0.0105142801966027\\
32	0.0105390095295132\\
33	0.0105803252460547\\
34	0.0106353566784881\\
35	0.0107004649885287\\
36	0.0107728933682781\\
37	0.0108544351069626\\
38	0.0109377588351573\\
39	0.0110202193214867\\
40	0.0110978897491019\\
41	0.0111743164199206\\
42	0.0112462491679149\\
43	0.011311790630699\\
44	0.0113721875066781\\
45	0.0114243122114231\\
46	0.0114732018148402\\
47	0.011512758893821\\
48	0.011546610240419\\
49	0.0115746936281492\\
50	0.0115949951955396\\
51	0.0116054531223218\\
52	0.0116069220595166\\
53	0.0115981295312465\\
54	0.0115761734317259\\
55	0.0115408865743737\\
56	0.0114890358892474\\
57	0.0114263537076534\\
58	0.0113522085463699\\
59	0.0112672030187448\\
60	0.0111703065236658\\
61	0.0110609560404214\\
62	0.0109395117161284\\
63	0.0108066591242513\\
64	0.0106693325394887\\
65	0.0105298010785002\\
66	0.0103880659481461\\
67	0.0102442188289175\\
68	0.0100991623389243\\
69	0.00995570363436403\\
70	0.00981600683746628\\
71	0.00967842065672312\\
72	0.00954397961236592\\
73	0.00941205240795842\\
74	0.0092822513563135\\
75	0.0091554738127625\\
76	0.00903335858504368\\
77	0.00891766471316289\\
78	0.00881033293459683\\
79	0.00871378533254674\\
80	0.00863013173305495\\
81	0.00856164349336867\\
82	0.00850759474105946\\
83	0.00845588344051854\\
84	0.00837570756078776\\
85	0.00824083873234511\\
86	0.0080550159623431\\
87	0.00783345832154744\\
88	0.00757507707514252\\
89	0.00729819420325366\\
90	0.00703151849026856\\
91	0.00679203428179121\\
92	0.00652591201056037\\
93	0.00622908240069871\\
94	0.005979702795125\\
95	0.00580687112200577\\
96	0.00565012715743107\\
97	0.00545771957034201\\
98	0.00524753993994346\\
99	0.0050672442479282\\
100	0.00495588115296243\\
100	-0.00499441785357724\\
99	-0.0049967681123404\\
98	-0.00497074958761609\\
97	-0.0049219808521166\\
96	-0.00491139323639931\\
95	-0.0050379784053656\\
94	-0.00532490931769906\\
93	-0.00576627382697927\\
92	-0.00633264685488404\\
91	-0.00697747687327624\\
90	-0.00767962948167699\\
89	-0.00851509671924815\\
88	-0.00945077987422481\\
87	-0.0104236960977366\\
86	-0.0114143177945962\\
85	-0.0124495756095539\\
84	-0.0135186903028732\\
83	-0.014457321892741\\
82	-0.0151127185929393\\
81	-0.0155094345913381\\
80	-0.0157561428459541\\
79	-0.015925571518345\\
78	-0.0160672686150816\\
77	-0.0162050166883898\\
76	-0.0163517886689373\\
75	-0.0165127724700454\\
74	-0.0166894424378321\\
73	-0.0168808492118192\\
72	-0.0170841454829834\\
71	-0.0172969120657756\\
70	-0.0175170635516107\\
69	-0.0177421370236838\\
68	-0.0179695832193899\\
67	-0.0181965953667687\\
66	-0.0184239636336782\\
65	-0.0186488111124329\\
64	-0.0188695503717738\\
63	-0.0190840675926207\\
62	-0.0192928902884034\\
61	-0.0194996834341507\\
60	-0.0197028821249\\
59	-0.019902555779865\\
58	-0.0200985109638553\\
57	-0.020291594907199\\
56	-0.0204830088275198\\
55	-0.0206719185815615\\
54	-0.0208655229452526\\
53	-0.0210618764967031\\
52	-0.021262005525993\\
51	-0.0214650237949744\\
50	-0.0216727811249506\\
49	-0.0218871711450258\\
48	-0.0221080467859822\\
47	-0.0223351212480566\\
46	-0.0225693996124794\\
45	-0.0228164795661328\\
44	-0.0230715254129007\\
43	-0.0233410082435579\\
42	-0.0236226876530636\\
41	-0.0239203606814283\\
40	-0.0242339512209834\\
39	-0.0245602734556468\\
38	-0.0249061787476041\\
37	-0.0252707247061274\\
36	-0.0256526356704172\\
35	-0.0260454790616563\\
34	-0.0264522858621931\\
33	-0.0268735759532363\\
32	-0.0273119622773123\\
31	-0.0277747986369979\\
30	-0.0282686792955442\\
29	-0.0287890540540034\\
28	-0.0293346728699458\\
27	-0.0298873250719925\\
26	-0.0304116129837184\\
25	-0.0308664639569698\\
24	-0.0311592106591566\\
23	-0.0311970182647147\\
22	-0.0309192069274263\\
21	-0.0304726788624062\\
20	-0.0301951655992033\\
19	-0.0310188712986363\\
18	-0.0352245930226811\\
17	-0.0429916959057171\\
16	-0.0535169681017651\\
15	-0.0646429514268561\\
14	-0.0703119443236652\\
13	-0.0688981977839201\\
12	-0.0695387701725597\\
11	-0.0753044968743224\\
10	-0.0820612319129916\\
9	-0.0864004239798862\\
8	-0.0891657912757413\\
7	-0.0908661622446804\\
6	-0.0919878153344315\\
5	-0.0927410838488349\\
4	-0.0931472664522293\\
3	-0.0930369216377712\\
2	-0.0912201269217389\\
1	-0.0957476650179134\\
0	-0.096\\
}--cycle;
\addplot [color=black,solid,forget plot]
  table[row sep=crcr]{%
0	0.1\\
1	0.099460360750204\\
2	0.103729221189449\\
3	0.101825826567739\\
4	0.101651338240887\\
5	0.101962630216452\\
6	0.102581710152676\\
7	0.103414838267884\\
8	0.10437747780429\\
9	0.105250107234207\\
10	0.104952276463737\\
11	0.10201988779293\\
12	0.091203945402814\\
13	0.0673934105497987\\
14	0.0334512104568801\\
15	0.00696513001828757\\
16	-0.00408914757939681\\
17	-0.00580685287946084\\
18	-0.00420699761961525\\
19	-0.0034783282209616\\
20	-0.004832418210829\\
21	-0.00653848551334927\\
22	-0.00794412945797805\\
23	-0.00893283998992854\\
24	-0.00948139369547747\\
25	-0.00969548934701984\\
26	-0.00968823772028905\\
27	-0.00955917535766935\\
28	-0.00935743477317032\\
29	-0.00912328641225445\\
30	-0.00887930738564158\\
31	-0.00863025922019762\\
32	-0.00838647637389956\\
33	-0.00814662535359081\\
34	-0.0079084645918525\\
35	-0.00767250703656377\\
36	-0.00743987115106953\\
37	-0.00720814479958242\\
38	-0.00698420995622339\\
39	-0.00677002706708007\\
40	-0.00656803073594072\\
41	-0.00637302213075385\\
42	-0.00618821924257433\\
43	-0.00601460880642944\\
44	-0.0058496689531113\\
45	-0.00569608367735486\\
46	-0.0055480988988196\\
47	-0.00541118117711782\\
48	-0.00528071827278161\\
49	-0.00515623875843834\\
50	-0.0050388929647055\\
51	-0.00492978533632631\\
52	-0.00482754173323822\\
53	-0.00473187348272829\\
54	-0.00464467475676333\\
55	-0.00456551600359391\\
56	-0.00449698646913619\\
57	-0.00443262059977282\\
58	-0.0043731512087427\\
59	-0.00431767638056011\\
60	-0.00426628780061714\\
61	-0.00421936369686463\\
62	-0.00417668928613751\\
63	-0.00413870423418466\\
64	-0.00410010891614255\\
65	-0.00405950501696635\\
66	-0.00401794884276603\\
67	-0.00397618826892563\\
68	-0.00393521044023276\\
69	-0.00389321669465991\\
70	-0.00385052835707222\\
71	-0.00380924570452622\\
72	-0.00377008293530876\\
73	-0.00373439840193039\\
74	-0.00370359554075932\\
75	-0.00367864932864144\\
76	-0.00365921504194682\\
77	-0.00364367598761347\\
78	-0.00362846784024239\\
79	-0.00360589309289913\\
80	-0.00356300555644956\\
81	-0.00347389554898472\\
82	-0.00330256192593991\\
83	-0.00300071922611122\\
84	-0.00257149137104271\\
85	-0.00210436843860438\\
86	-0.00167965091612655\\
87	-0.00129511888809458\\
88	-0.000937851399541146\\
89	-0.000608451257997243\\
90	-0.000324055495704212\\
91	-9.27212957425152e-05\\
92	9.66325778381634e-05\\
93	0.00023140428685972\\
94	0.000327396738712972\\
95	0.000384446358320083\\
96	0.000369366960515878\\
97	0.000267869359112708\\
98	0.000138395176163684\\
99	3.52380677938978e-05\\
100	-1.92683503074042e-05\\
};
\addplot [color=black,dashed,forget plot]
  table[row sep=crcr]{%
0	0\\
100	0\\
};
\end{axis}
\end{tikzpicture}%
}
  \vspace*{10pt}
  \pgfplotsset{axis on top, ytick align=outside, xtick align=outside, xlabel={Time (s)}, xmin=0, xmax=65}
%
%
\begin{tikzpicture}

\begin{axis}[%
width=0.951\figurewidth,
height=\figureheight,
at={(0\figurewidth,0\figureheight)},
scale only axis,
xmin=0,
ymin=-0.5,
ymax=0.5,
ylabel={$k_2$},
axis background/.style={fill=white},
axis x line*=bottom,
axis y line*=left
]

\addplot[area legend,solid,draw=white!70!black,fill=white!70!black,forget plot]
table[row sep=crcr] {%
x	y\\
0	0.296\\
1	0.295996151790116\\
2	0.296015962542549\\
3	0.296001588567795\\
4	0.296003869077795\\
5	0.296011722784037\\
6	0.29602438138903\\
7	0.296040091500723\\
8	0.296050225157841\\
9	0.296040880872007\\
10	0.295982038322468\\
11	0.295853112041929\\
12	0.295478355634155\\
13	0.294529978759611\\
14	0.292353566467353\\
15	0.2878286647568\\
16	0.279116599552286\\
17	0.26534155928755\\
18	0.25094294305006\\
19	0.242601381477129\\
20	0.239174499871145\\
21	0.239840292637745\\
22	0.244131106014805\\
23	0.249085276273135\\
24	0.252166534244833\\
25	0.252984580850773\\
26	0.252064134853109\\
27	0.250146854900953\\
28	0.247728200374269\\
29	0.245157560537739\\
30	0.242624756519529\\
31	0.240193155010927\\
32	0.237916671766993\\
33	0.235777895408259\\
34	0.233749171345461\\
35	0.231825486311103\\
36	0.230004507098547\\
37	0.228264141798331\\
38	0.226629116005997\\
39	0.22509563657173\\
40	0.223667759841968\\
41	0.2223084085689\\
42	0.221026449048577\\
43	0.219822417596972\\
44	0.21867407603625\\
45	0.217595155008921\\
46	0.216544235551895\\
47	0.215552082092873\\
48	0.214586243278008\\
49	0.21364006374037\\
50	0.212719014226901\\
51	0.211826591159648\\
52	0.210952029786484\\
53	0.210087686025865\\
54	0.209242609814072\\
55	0.208407905318665\\
56	0.207598919755806\\
57	0.206772562475087\\
58	0.205934739634077\\
59	0.205080089957279\\
60	0.204209074411734\\
61	0.203324484892129\\
62	0.202428156981141\\
63	0.201529402416303\\
64	0.200605629843261\\
65	0.199655523469198\\
66	0.198690678801599\\
67	0.197720204762618\\
68	0.196756594531755\\
69	0.19579385842082\\
70	0.194839746922059\\
71	0.193902655167112\\
72	0.192986210460714\\
73	0.192085773151447\\
74	0.191186378140426\\
75	0.190263815102692\\
76	0.189249145890906\\
77	0.188028111046211\\
78	0.186409115531253\\
79	0.184051757442908\\
80	0.180460952035221\\
81	0.174698188074317\\
82	0.165686209246465\\
83	0.151901343695231\\
84	0.133626602522751\\
85	0.113965422045282\\
86	0.095677997726414\\
87	0.0788441761985985\\
88	0.0630224058266466\\
89	0.0486540976449758\\
90	0.0368650677333747\\
91	0.0282018815931949\\
92	0.0216346049430239\\
93	0.0169321788752451\\
94	0.014061009407712\\
95	0.0126472792895039\\
96	0.0120842704825634\\
97	0.0120647906311815\\
98	0.0122299921751527\\
99	0.0123718485598266\\
100	0.0124256723242623\\
100	-0.0106478837143123\\
99	-0.0109538265152109\\
98	-0.0114213268210763\\
97	-0.0119911573528588\\
96	-0.0126430131074911\\
95	-0.0134424515144068\\
94	-0.0146287516531275\\
93	-0.0167156035159089\\
92	-0.0197896753383796\\
91	-0.0235376500653844\\
90	-0.0280456649495448\\
89	-0.0338962480372895\\
88	-0.040548574036665\\
87	-0.0472188568971103\\
86	-0.0534849205864722\\
85	-0.0593940830677754\\
84	-0.0646114482728109\\
83	-0.0684992370882024\\
82	-0.0708882937626138\\
81	-0.0722275655300626\\
80	-0.0730104786243751\\
79	-0.0734573204970231\\
78	-0.073732055366108\\
77	-0.0738975350536941\\
76	-0.0740030686641188\\
75	-0.0740824125501521\\
74	-0.0741516792405029\\
73	-0.0742210886926351\\
72	-0.0743048529734137\\
71	-0.0744061392007629\\
70	-0.074525581502289\\
69	-0.0746651831620018\\
68	-0.0748305632364338\\
67	-0.0750218565364579\\
66	-0.0752278231508484\\
65	-0.0754496508302387\\
64	-0.0756882160974173\\
63	-0.0759471058551827\\
62	-0.0762170712644626\\
61	-0.0764628609144098\\
60	-0.076684091432645\\
59	-0.0768768709460642\\
58	-0.0770419921904045\\
57	-0.077182451262997\\
56	-0.0772920991830287\\
55	-0.0773737108942795\\
54	-0.0773850179332033\\
53	-0.0773451479012203\\
52	-0.0772494324152872\\
51	-0.0771142312662622\\
50	-0.0769430126805925\\
49	-0.0767279079164316\\
48	-0.0764802478933195\\
47	-0.0762121648498626\\
46	-0.0759207166080413\\
45	-0.0755796460321422\\
44	-0.0752271826118376\\
43	-0.0748254598400902\\
42	-0.0743929946751725\\
41	-0.0739168775062131\\
40	-0.0734056524272422\\
39	-0.0728707844341662\\
38	-0.072285967064538\\
37	-0.0716681049335797\\
36	-0.0710204233759234\\
35	-0.0703747682736059\\
34	-0.0697138533884196\\
33	-0.0690489503803286\\
32	-0.0683973374147317\\
31	-0.0677638310001876\\
30	-0.067159644000843\\
29	-0.0666679532748873\\
28	-0.0663612624896219\\
27	-0.0664352176750116\\
26	-0.0672318800724058\\
25	-0.0692139006444934\\
24	-0.0730579454522467\\
23	-0.0791904534361907\\
22	-0.0871019643199089\\
21	-0.0948130336114316\\
20	-0.101053301544415\\
19	-0.105977498800313\\
18	-0.109421414014544\\
17	-0.108579452270268\\
16	-0.1045021055608\\
15	-0.100698068081888\\
14	-0.098266973736696\\
13	-0.0968989560957677\\
12	-0.0962846960421374\\
11	-0.0960563438143209\\
10	-0.0959873788091565\\
9	-0.0959498612271111\\
8	-0.0959470769399006\\
7	-0.0959591431900616\\
6	-0.0959753494820493\\
5	-0.0959881217618715\\
4	-0.0959960164729961\\
3	-0.0959983194125272\\
2	-0.0959839764700584\\
1	-0.0960038136145781\\
0	-0.096\\
}--cycle;
\addplot [color=black,solid,forget plot]
  table[row sep=crcr]{%
0	0.1\\
1	0.099996169087769\\
2	0.100015993036245\\
3	0.100001634577634\\
4	0.100003926302399\\
5	0.100011800511083\\
6	0.10002451595349\\
7	0.100040474155331\\
8	0.10005157410897\\
9	0.100045509822448\\
10	0.0999973297566559\\
11	0.099898384113804\\
12	0.0995968297960085\\
13	0.0988155113319215\\
14	0.0970432963653287\\
15	0.0935652983374563\\
16	0.0873072469957429\\
17	0.0783810535086411\\
18	0.0707607645177577\\
19	0.068311941338408\\
20	0.0690605991633646\\
21	0.0725136295131565\\
22	0.0785145708474479\\
23	0.0849474114184721\\
24	0.089554294396293\\
25	0.0918853401031398\\
26	0.0924161273903514\\
27	0.0918558186129706\\
28	0.0906834689423236\\
29	0.0892448036314257\\
30	0.0877325562593432\\
31	0.0862146620053696\\
32	0.0847596671761305\\
33	0.0833644725139652\\
34	0.0820176589785208\\
35	0.0807253590187487\\
36	0.0794920418613117\\
37	0.0782980184323755\\
38	0.0771715744707294\\
39	0.076112426068782\\
40	0.0751310537073627\\
41	0.0741957655313434\\
42	0.0733167271867023\\
43	0.072498478878441\\
44	0.0717234467122063\\
45	0.0710077544883896\\
46	0.0703117594719271\\
47	0.0696699586215051\\
48	0.069052997692344\\
49	0.0684560779119694\\
50	0.067888000773154\\
51	0.0673561799466928\\
52	0.0668512986855984\\
53	0.0663712690623223\\
54	0.0659287959404345\\
55	0.065517097212193\\
56	0.0651534102863886\\
57	0.0647950556060449\\
58	0.064446373721836\\
59	0.0641016095056073\\
60	0.0637624914895443\\
61	0.0634308119888598\\
62	0.0631055428583391\\
63	0.0627911482805599\\
64	0.0624587068729218\\
65	0.0621029363194795\\
66	0.0617314278253753\\
67	0.06134917411308\\
68	0.0609630156476604\\
69	0.0605643376294092\\
70	0.0601570827098847\\
71	0.0597482579831746\\
72	0.0593406787436501\\
73	0.0589323422294059\\
74	0.0585173494499615\\
75	0.0580907012762699\\
76	0.0576230386133936\\
77	0.0570652879962582\\
78	0.0563385300825723\\
79	0.0552972184729423\\
80	0.053725236705423\\
81	0.051235311272127\\
82	0.0473989577419258\\
83	0.0417010533035144\\
84	0.0345075771249702\\
85	0.0272856694887532\\
86	0.0210965385699709\\
87	0.0158126596507441\\
88	0.0112369158949908\\
89	0.00737892480384314\\
90	0.00440970139191494\\
91	0.00233211576390522\\
92	0.000922464802322194\\
93	0.00010828767966808\\
94	-0.000283871122707734\\
95	-0.00039758611245145\\
96	-0.000279371312463844\\
97	3.68166391613507e-05\\
98	0.000404332677038203\\
99	0.000709011022307869\\
100	0.000888894304974977\\
};
\addplot [color=black,dashed,forget plot]
  table[row sep=crcr]{%
0	0\\
100	0\\
};
\end{axis}
\end{tikzpicture}%
  \caption{Evolution of the camera calibration when started from distant initial values in one Monte Carlo simulation. Initial convergence is reached immediately after the first image pair has been observed, and final convergence once the camera has moved sufficiently. The shaded regions depict the 95\% uncertainty interval given by the state-space estimation.}
  \label{fig:evolution}
\end{figure}

To analyze the observability of the parameters in the simulated data, we provide results also for a bundle-adjusted batch solution (`Batch opt.' in Table~\ref{tbl:results}), which acts as a brute-force baseline for the methods. For this baseline the camera parameters and motion track estimated by our proposed method were used for initializing the bundle adjustment (non-linear minimization of the reprojection error). The bundle adjustment problem was solved by constrained minimization using the \textsc{Matlab} \texttt{fmincon} Interior Point Algorithm. This bundle adjustment approach was computationally very intensive, but the results show that the data is informative about the parameter values, as the optimization converges to the ground-truth values.

The third method was the method by Jia and Evans \cite{Jia+Evans:2014}. We used their autocalibration toolbox\footnote{Online Camera-Gyroscope Auto-Calibration for Cellphones: \\ \url{http://users.ece.utexas.edu/~bevans/papers/2015/autocalibration/}} implementation for solving the self-calibration problem. As can be seen in the results, there were issues in running these results. As the method by Jia and Evans assumes there to be a lot of features available, we re-ran the results by including almost ten time more features in the simulation. The method keeps oscillating around the correct camera parameter values, but fails to converge. We suspect that the issues are partly related to the use case not being optimal for their method, and partly because of issues in their implementation of the method.

Figure~\ref{fig:evolution} shows the evolution of the camera calibration, when started from distant initial values. The shaded regions depict the 95\% uncertainty interval given by the state-space estimation. The model reaches initial convergence immediately after the first image pair has been observed, and final convergence once the camera has moved sufficiently.Note that $k_2$ does not show the convergence of the other parameters. The main reason is that for this case the features are around the center of the image at the beginning, where the $r$ from \ref{eq:radial2} is small and thus the gradient with respect to $k_2$ is small. For the synthetic case, the distortion parameters are an over-model.

The results show that the proposed method has an RMSE less than one in pixel units for the intrinsic parameters, not far from the results given brute-force calculated bundle-adjustment based calibration results. The method by Jia and Evans is clearly off more, but still did not diverge.

\begin{figure}[!t]
  \centering\footnotesize

  \setlength{\figurewidth}{.22\columnwidth}
  \setlength{\figureheight}{0.75\figurewidth}
  \hspace*{\fill}
  \includegraphics[width=\figurewidth]{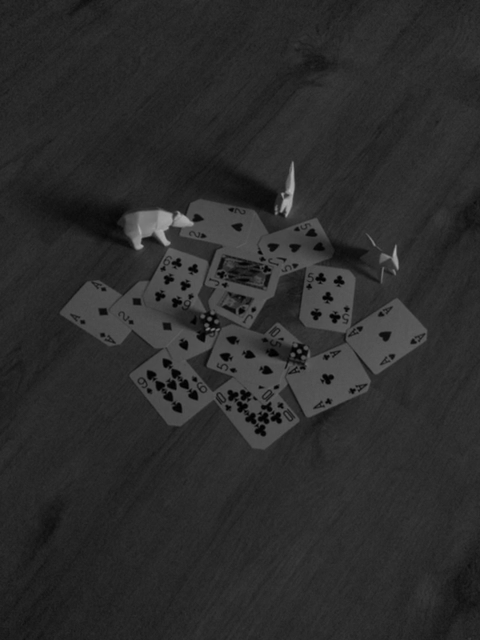} \hspace*{\fill}
  \includegraphics[width=\figurewidth]{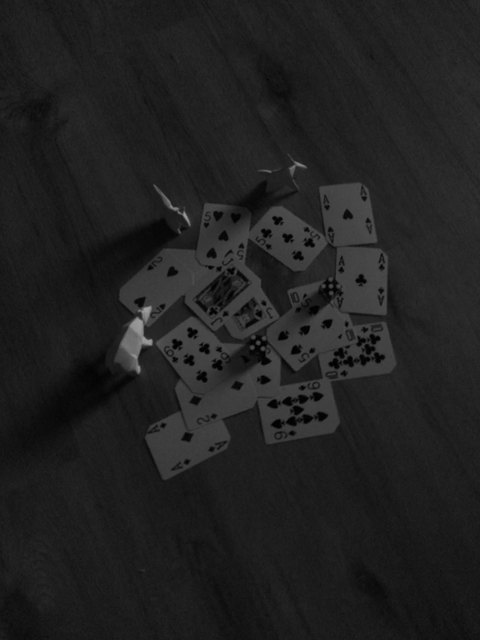} \hspace*{\fill}
  \includegraphics[width=\figurewidth]{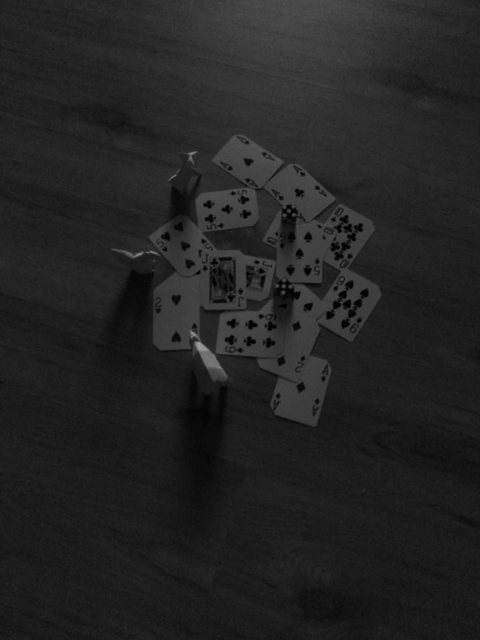} \hspace*{\fill}
  \includegraphics[width=\figurewidth]{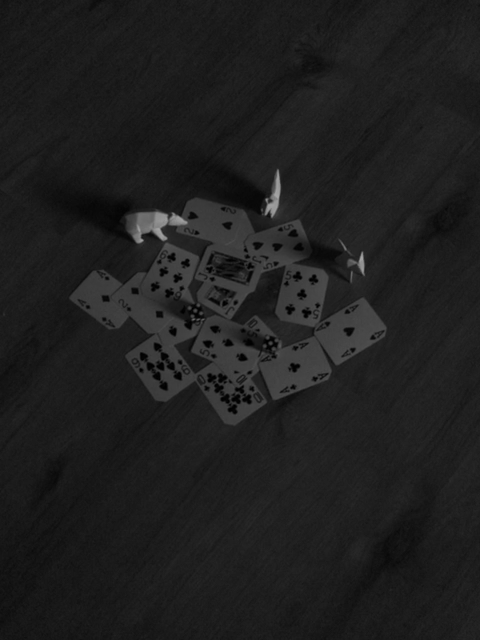} \hspace*{\fill} \\
  (a)~Cards---A dark indoor data set \\[1em]
  \hspace*{\fill}
  \includegraphics[width=\figurewidth]{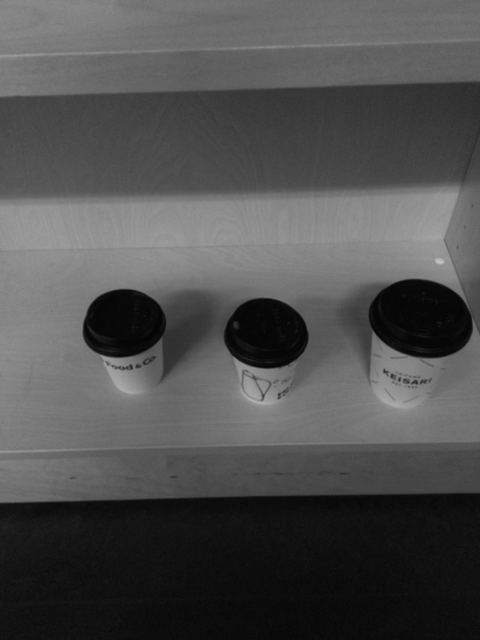} \hspace*{\fill}
  \includegraphics[width=\figurewidth]{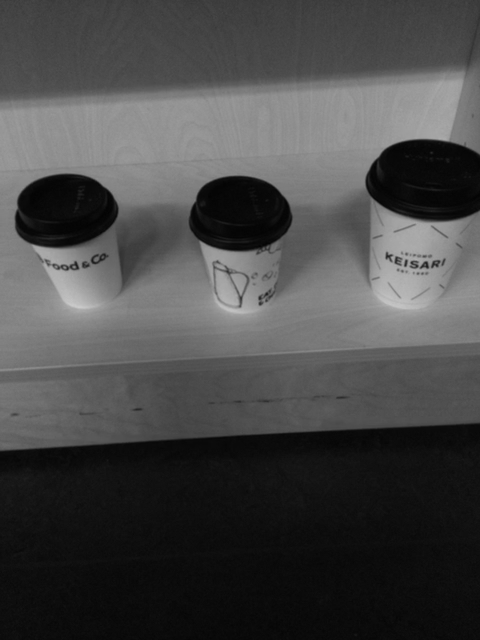} \hspace*{\fill}
  \includegraphics[width=\figurewidth]{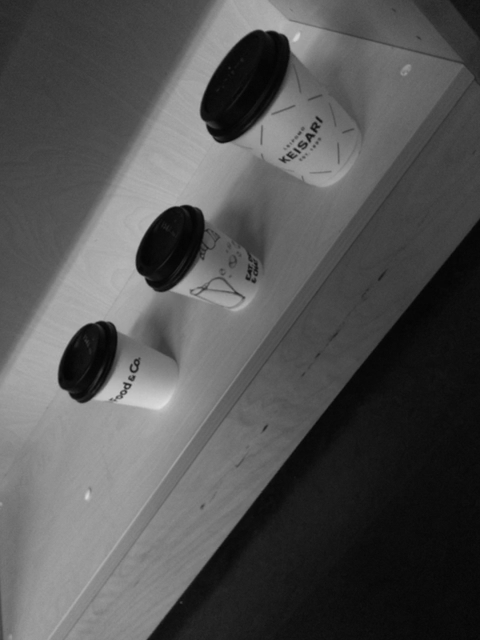} \hspace*{\fill}
  \includegraphics[width=\figurewidth]{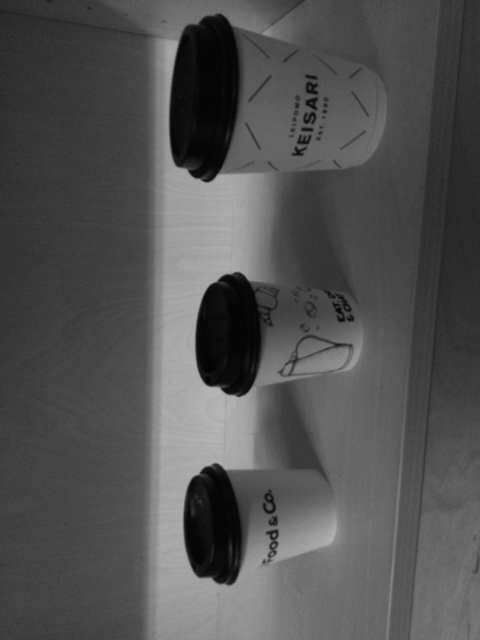} \hspace*{\fill} \\
  (b)~Cups---A feature-poor data set \\[1em]
  \hspace*{\fill}
  \includegraphics[width=\figurewidth]{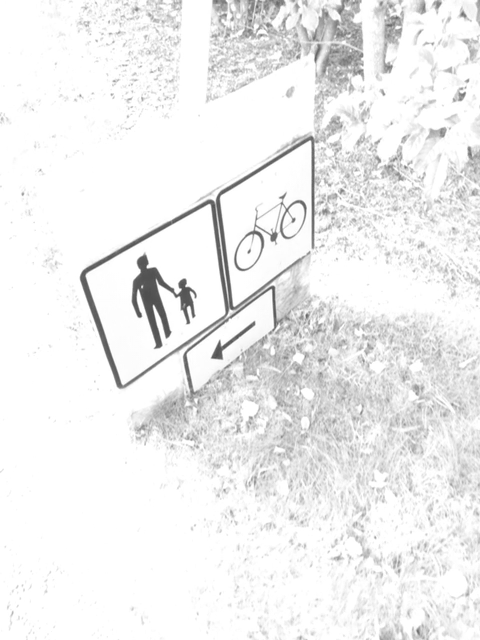} \hspace*{\fill}
  \includegraphics[width=\figurewidth]{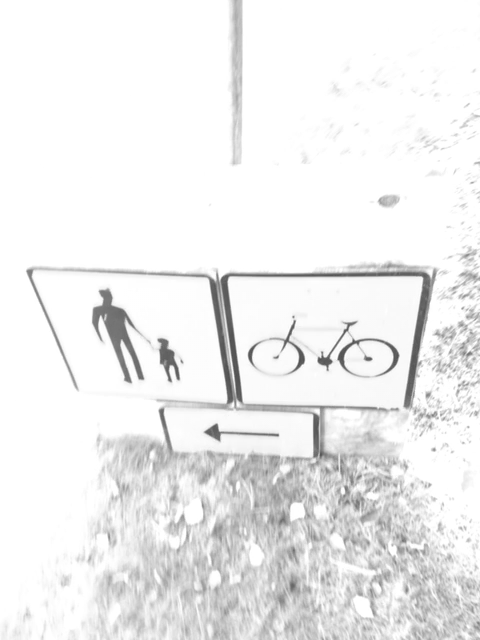} \hspace*{\fill}
  \includegraphics[width=\figurewidth]{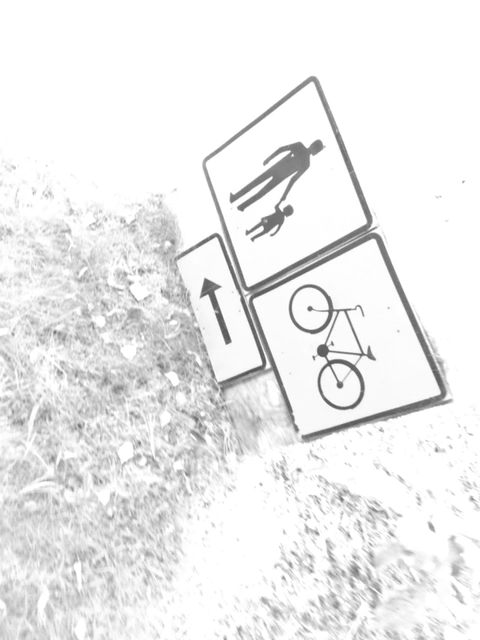} \hspace*{\fill}
  \includegraphics[width=\figurewidth]{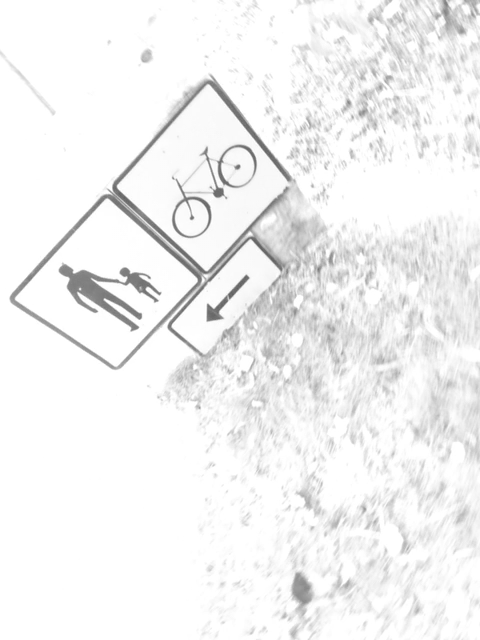} \hspace*{\fill} \\
  (c)~Signs---Overexposed outdoor data set \\[1em]
  \caption{Example frames from the empirical tests: (a)~is underexposed and with uneven concentration of features, (b)~is from a feature-poor indoor scene, and (c)~is an overexposed outdoor scene.}
  \label{fig:frames}
\end{figure}

\begin{table}
  \caption{Parameter values in empirical experiments.}
  \label{tbl:emp_results}
  \centering\footnotesize
  \begin{tabular}{ l c c c c c c c  } 
  \toprule
  Variable & Initial & Final & Ground-truth\\
  \midrule
  $f_x$ (px) & 	700 & 	568.12 & 570.56\\
  $f_y$ (px) & 	700& 	570.64&  569.72\\
  $c_x$ (px) & 	240 & 	238.11& 	238.03 \\
  $c_y$ (px) & 	320 & 	325.09& 	323.87 \\
  $k_1$  &     	0& 	0.0429 &	 	0.1134 \\
  $k_2$  &     	0 & 	$-0.0040$ & $-0.0634$ \\
  \bottomrule
  \end{tabular}
\end{table}

\subsection{Empirical Tests}
\noindent
In order to demonstrate the proposed online calibration approach on empirical data, we acquired test data in various situations using an Apple iPad Pro 12.9-inch model. Hardware-wise the iPad can be seen as a representative example of a modern-day smart-device. Similar sensors are available in most Andoroid and iOS smartphones and tablets.

The data acquisition was conducted using a custom data capture app implemented in Objective-C. The capture tool app stored the three-axis gyroscope together with associated timestamps to a file in the device. The gyroscope sampling rate was set to 100~Hz. The gyroscope was pre-calibrated before the data acquisition for estimating the additive gyroscope bias $\vectb{\omega}_b$.

Simultaneously to the gyroscope capture, device camera frames were read time-locked to the gyroscope observations. The experiments use the rear-facing camera with a resolution of $480 \times 640$ (portrait orientation), grayscale images, exposure time $1/60$, fixed aperture, sensitivity (ISO value) $125$, and locked focus at infinity. The camera refresh rate was 10~fps (Hz). The camera frames were stored as H.264 packed video sequences on the device, with exact frame timestamps stored separately for use in the offline run.

Furthermore, camera images of the canonical OpenCV checkerboard pattern were captured for conducting  batch calibration of the tablet camera. This calibration was only used for obtaining ground-truth camera parameters to compare against.

For obtaining a set of verstaile use cases, we recorded short sequences of a few controlled static scenes. In these data sets the motion of the camera was smooth in the sense of avoiding hard stops, in order to comply with the constrains of the model. Figure~\ref{fig:frames} shows example frames from the empirical data sets. They include (a)~ill-lit (underexposed) indoor scenes (we name this data set `cards'), (b)~visual feature poor scenes (typical in office environments), and (c)~overexposed outdoor scenes. 

\begin{figure}[!t]
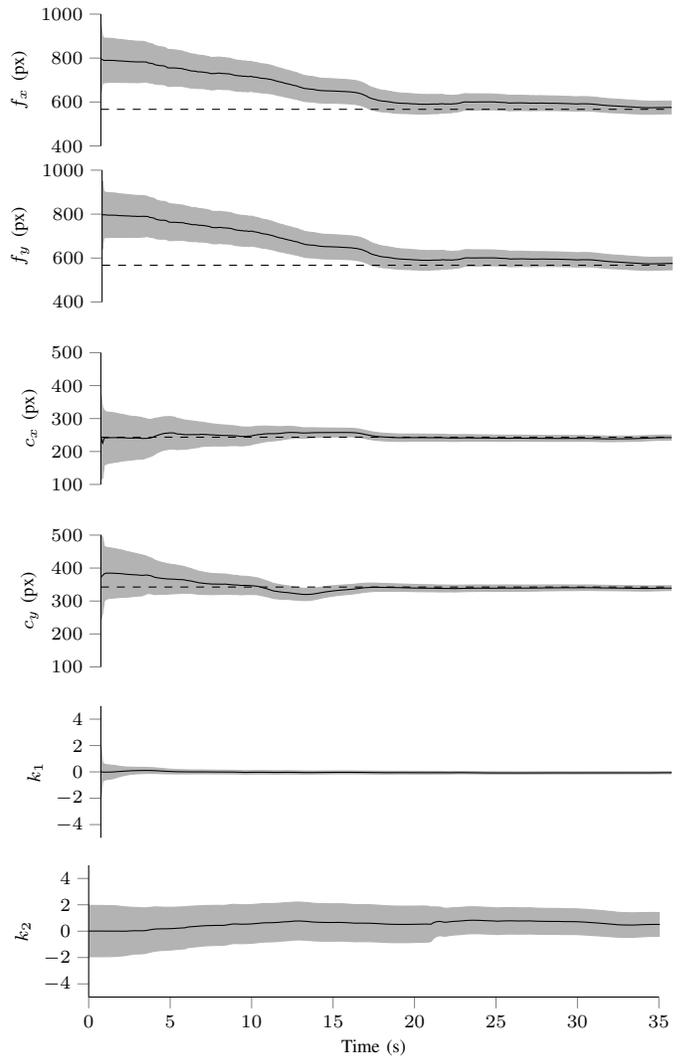


  \setlength{\figurewidth}{.90\columnwidth}
  \setlength{\figureheight}{0.22\figurewidth}
  \raggedleft\scriptsize%
  {\pgfplotsset{axis on top, hide x axis, ytick align=outside, xmin=0, xmax=35}
  \input{./fig/real_focal_x.tex}
  \vspace*{10pt}
  \input{./fig/real_focal_y.tex}
  \vspace*{10pt}
  \input{./fig/real_cent_x.tex}
  \vspace*{10pt}
  \input{./fig/real_cent_y.tex}
  \vspace*{10pt}
  \input{./fig/real_rad_1.tex}}
  \vspace*{10pt}  
  \pgfplotsset{axis on top, ytick align=outside, xtick align=outside, xlabel={Time (s)}, xmin=0, xmax=35}
  \input{./fig/real_rad_2.tex}
  \caption{Evolution of the camera calibration in the `Cards' data set, when started from distant initial values. The shaded regions depict the 95\% uncertainty interval given by the state-space estimation. The convergence occurs once sufficient excitation movement has occurred.}
  \label{fig:evolution_empirical}
\end{figure}

The evolution of the proposed calibration method is shown in Figure~\ref{fig:evolution_empirical}, where the shaded area represents the 95\% uncertainty interval. The corresponding calibration result is shown in Table~\ref{tbl:emp_results} together with the initial values and the checkerboard calibrated ground-truth for the device camera.

\begin{figure*}[!t]

  \setlength{\figurewidth}{\textwidth}
  \setlength{\figureheight}{0.22\figurewidth}
  \pgfplotsset{axis on top, y tick label style={rotate=90}, legend columns=-1}
  \centering\small
  \begin{subfigure}[b]{\textwidth}
    \raggedleft\footnotesize
    \input{./fig/position.tex}
    \caption{Latent camera position state estimates}
  \end{subfigure}
  \\[2pt]
  \begin{subfigure}[b]{\textwidth}
    \raggedleft\footnotesize
    \input{./fig/orientation.tex}
    \caption{Camera orientation estimates}
  \end{subfigure}
  \\[1em]
  \begin{subfigure}[b]{\textwidth}
    \centering\footnotesize
    \setlength{\figurewidth}{.09\textwidth}
    \hspace*{\fill}
    \includegraphics[width=\figurewidth]{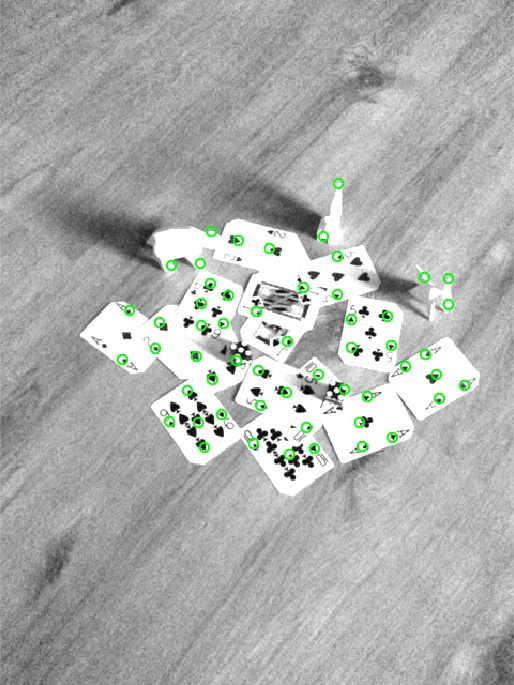}
    \hspace*{\fill}
    \includegraphics[width=\figurewidth]{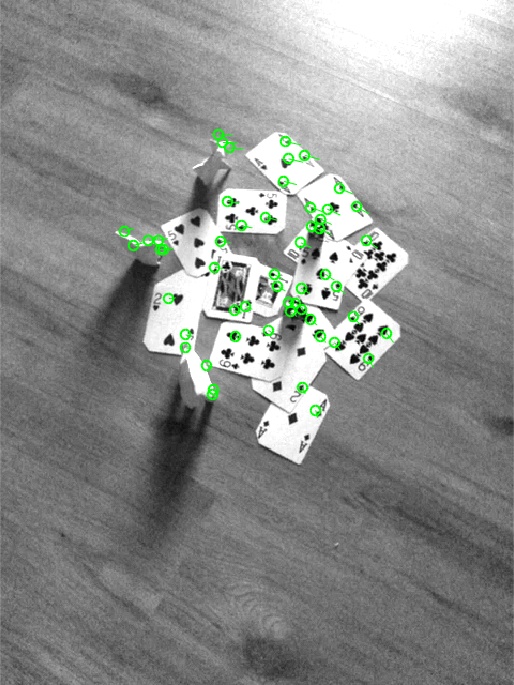}
    \hspace*{\fill}
    \includegraphics[width=\figurewidth]{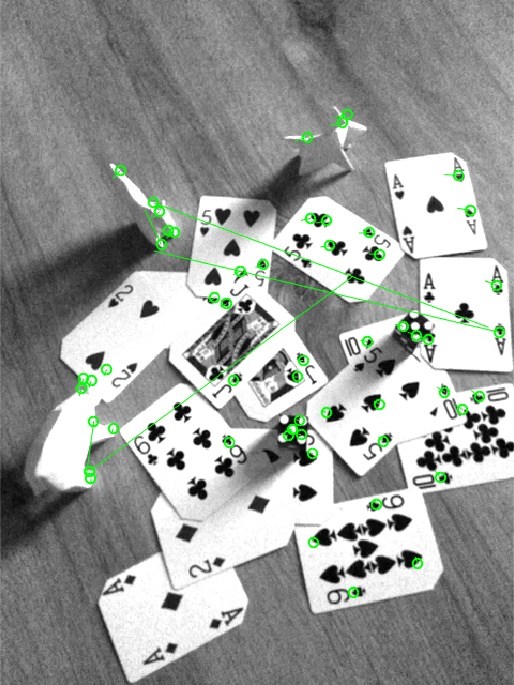}
    \hspace*{\fill}
    \includegraphics[width=\figurewidth]{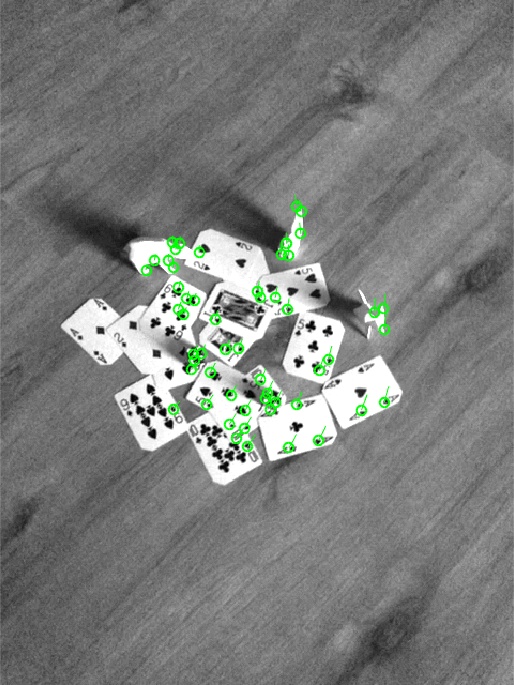}
    \hspace*{\fill}
    \includegraphics[width=\figurewidth]{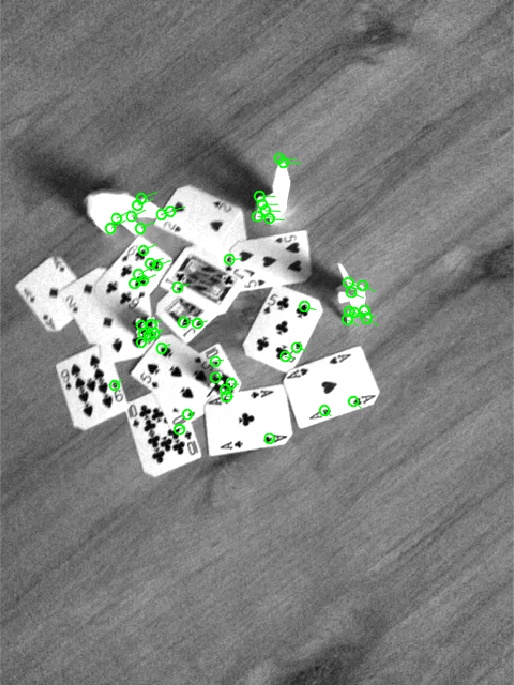}
    \hspace*{\fill}
    \includegraphics[width=\figurewidth]{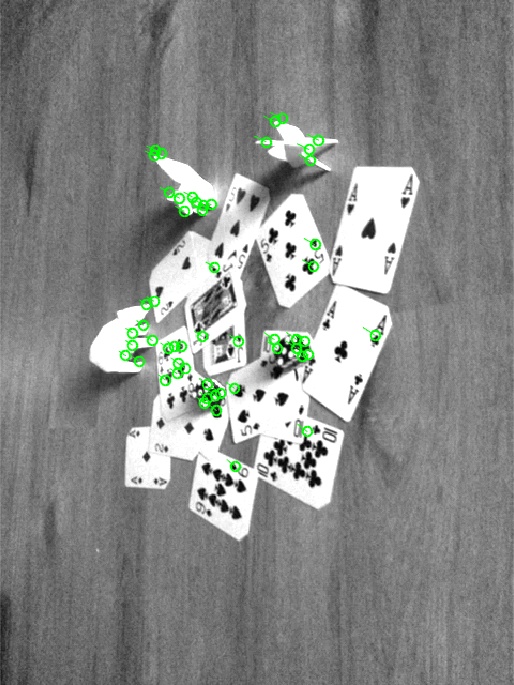}
    \hspace*{\fill}
    \includegraphics[width=\figurewidth]{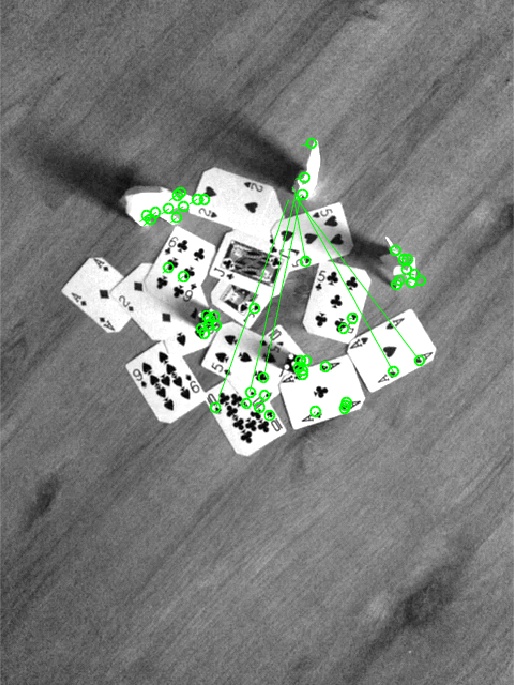}
    \hspace*{\fill}
    \includegraphics[width=\figurewidth]{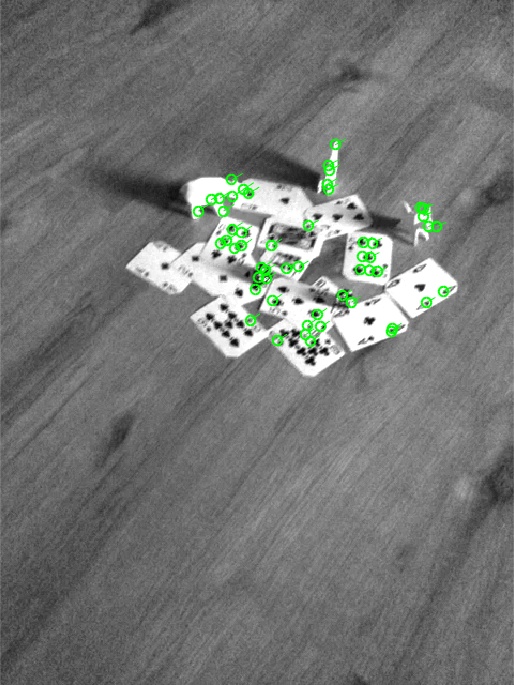}
    \hspace*{\fill}
    \includegraphics[width=\figurewidth]{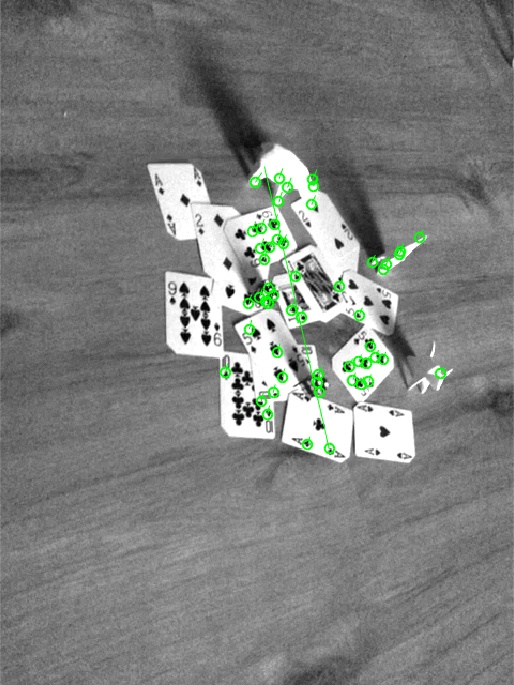}
    \hspace*{\fill}    
    \includegraphics[width=\figurewidth]{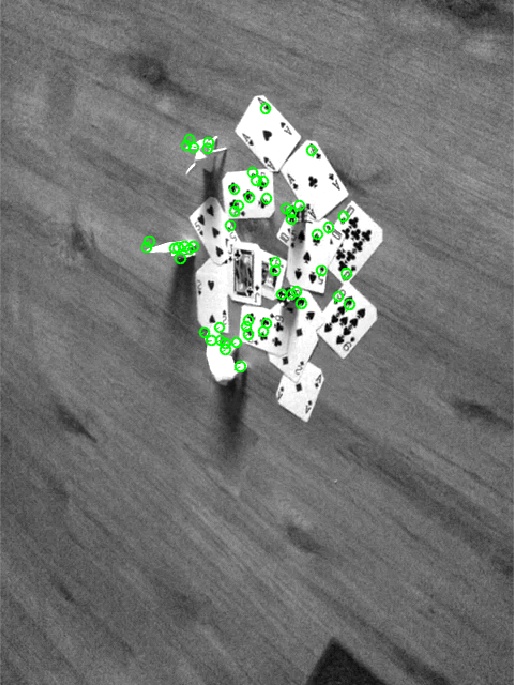}
    \hspace*{\fill}
    \\[6pt]
    \hspace*{\fill}
    \parbox[c]{\figurewidth}{\centering $t=0~\text{s}$}
    \hspace*{\fill}
    \parbox[c]{\figurewidth}{\centering $t=5~\text{s}$}
    \hspace*{\fill}
    \parbox[c]{\figurewidth}{\centering $t=10~\text{s}$}
    \hspace*{\fill}
    \parbox[c]{\figurewidth}{\centering $t=15~\text{s}$}
    \hspace*{\fill}    
    \parbox[c]{\figurewidth}{\centering $t=20~\text{s}$}
    \hspace*{\fill}
    \parbox[c]{\figurewidth}{\centering $t=25~\text{s}$}
    \hspace*{\fill}
    \parbox[c]{\figurewidth}{\centering $t=30~\text{s}$}
    \hspace*{\fill}
    \parbox[c]{\figurewidth}{\centering $t=35~\text{s}$}
    \hspace*{\fill}
    \parbox[c]{\figurewidth}{\centering $t=40~\text{s}$}
    \hspace*{\fill}
    \parbox[c]{\figurewidth}{\centering $t=45~\text{s}$}
    \hspace*{\fill}
    
    \caption{Associated frames at different time points}
  \end{subfigure}
  \caption{State estimates and example input frames for the parameter estimate track results in Fig.~\ref{fig:evolution_empirical}. Subfigure~(a) shows the latent (unobserved) camera position track estimates (the shaded area represents the 95\% credible interval) over the time-span of the data. Note that the absolute scale of the movement remains unobserved. (b)~The corresponding orientation states, which coincide more clearly with the behavior in Fig.~\ref{fig:evolution_empirical}. (c)~Example input frames in the cards data set with green markers showing the tracked feature positions. Observation times correspond to the tick marks in the above plots.}
  \label{fig:position-orientation}
\end{figure*}

For this run, the evolution of the position estimates and the orientation (transformed into Euler angles) is visualized in Figure~\ref{fig:position-orientation}. Subfigure~(a) shows the latent (unobserved) camera position track estimates (the shaded area represents the 95\% credible interval) over the time-span of the data. The absolute scale of the movement remains unobserved, which is due to the gyroscope only observing the rotation rate and the camera data being agnostic to the true distance of any feature movement. Subfigure~(b) shows the corresponding orientation states, which coincide more clearly with the behavior in Figure~\ref{fig:evolution_empirical}. The linear parameters appear to reach the right regime after the camera has been rotated sufficiently (\ie\ the data features sufficient excitation). The non-linear distortion parameters take longer to stabilize. 

Subfigure~\ref{fig:position-orientation}(c) shows the corresponding input frames in the cards data set with challenging lighting conditions. The green markers show the current feature point locations in the frames, and their `tails' (green line) show the point movement since the previous frame (frames sampled at 10~Hz).

For the intrinsic parameters, the estimated parameter values converge within a few pixels to the checkerboard-calibrated ground-truth values. For the non-linear radial distortion parameters the values are also similar. In case of the distortion parameters, the identification might suffer from the low feature concentration towards the edges of the visual frame data.

\section{Discussion and Conclusion}
\noindent
In this paper we have proposed a method for estimating the intrinsic parameters and lens distortion coefficients for camera calibration. This paper proposed a model for estimating the parameters on the fly by fusing gyroscope and camera data, where the model is based on joint estimation of visual feature positions, camera parameters, and the camera pose, the rotation of which is assumed to follow the movement predicted by the gyroscope. 

The estimation procedure is lightweight and performs online using an extended Kalman filter. The strengths of the method is in robustness to feature-poor visual environments and insensitivity to initialization, the aspects which were also demonstrated in the experiments on simulated and empirical data.

The experiments showed that the method performs well against the current state-of-the-art in gyroscope-aided self-calibration. The results showed that after sufficient motion the convergence is both quick, and it is easy to capture the moment of convergence by monitoring the state variance estimates for the camera parameters.

The empirical tests demonstrated the method using data captured using an Apple iPad Pro. The empirical data the parameter estimates converged rapidly after the system had experienced sufficient motion required for jointly estimating both the camera poses and feature world coordinates.

Figure~\ref{fig:frames} shows three distinctive data sets; one is underexposed and with uneven concentration of features, one is from a feature-poor indoor scene, and the third is from an overexposed outdoor scene. The proposed method is not sensitive to feature-poor environments as it can rely on a relatively low number of features being tracked---in the simulations only 27 features were used.

This is a clear difference to previous methods, which have been mostly inspired by building on purely machine vision methods. Requiring hundreds of tracked features to be available works well in controlled environments, in good lighting conditions, and using high-quality camera hardware. Our method focuses on the opposite---low quality data and a low number of features suffices. On the other hand, this comes with a requirement of the features remaining visible for a sufficiently long time period.

The method could be extended to include further parameters. Natural directions of extension would be to also estimate the additive three-axis gyroscope bias, or include further camera parameters such as rolling-shutter timing parameters.

As the method jointly estimates both camera poses, parameters, and feature positions, it can be sensitive to certain use cases and environments. For example, in Figure~\ref{fig:frames}(b) the cups are round and their corresponding features are often lost and picked up again during movement. Short feature tracks inject uncertainty into the estimation scheme in the proposed method, which slows down convergence and misleads the estimation. This could be improved by tuning the method parameters, or introducing visual loop-closures which would reuse the same features when they appear again.

Code implementing the method is available online: \\ \mbox{\url{https://aaltovision.github.io/camera-gyro-calibration/}}

\section*{Acknowledgments}
\noindent
This research was supported by the Academy of Finland grants 308640,
277685, and 295081.

{\small
\bibliographystyle{ieee}
\bibliography{bibliography}
}

\end{document}